\documentclass[letterpaper, 10 pt, conference]{ieeeconf}

\usepackage{silence}
\usepackage{graphicx}
\usepackage{xcolor,balance}
\usepackage{amsmath}
\usepackage{amssymb}

\usepackage{subcaption}
\usepackage[font=footnotesize]{caption} %

\usepackage{url}

\usepackage[sort,compress]{cite}
\makeatletter
\let\NAT@parse\undefined
\makeatother
\usepackage[
   colorlinks=true,
   linkcolor=black,
   citecolor=black,
   urlcolor=black,
   filecolor=black,
   pagebackref=false,
   hypertexnames=false,
   plainpages=false
]{hyperref}

\usepackage{bm}

\let\labelindent\relax
\usepackage{enumitem}

\usepackage[normalem]{ulem}

\usepackage{multirow}
\usepackage{multicol}
\usepackage{colortbl}

\usepackage{algorithm}
\usepackage{algpseudocode} 

\usepackage{pifont}

\usepackage{mathtools}

\usepackage{makecell}

\usepackage{booktabs}

\usepackage{tabularx}

\usepackage{arydshln}

\usepackage{threeparttable}

\usepackage{cuted}

\usepackage{lmodern}

\usepackage{courier} 

\usepackage{xspace}

\usepackage{tikz}

\usepackage{mathtools} %

\newcommand{\fields}{\mathbb{F}}

\newcommand{\SE}{\operatorname{SE}}

\definecolor{Red}{RGB}{200,30,30}     %
\definecolor{Green}{RGB}{30,130,40}     %
\definecolor{Blue}{RGB}{30,80,200}     %
\definecolor{Orange}{RGB}{210,110,20}    %
\definecolor{Brown}{RGB}{130,70,30}      %
\definecolor{Purple}{RGB}{130,50,170}    %

\definecolor{GreenBG}{RGB}{220,255,220}    %

\newcommand{\Iz}{{\textcolor{Blue}{\mathbf{I}_0}}}
\newcommand{\vz}{{\textcolor{Blue}{\mathbf{v}_0}}}

\newcommand{\Ii}{{\textcolor{Red}{\mathbf{I}_i}}}
\newcommand{\vi}{{\textcolor{Red}{\mathbf{v}_i}}}

\newcommand{\vzhat}{\textcolor{Blue}{\mathbf{\hat{v}}_0}}
\newcommand{\vihat}{\textcolor{Red}{\mathbf{\hat{v}}_i}}

\newcommand{\vzbar}{\textcolor{Blue}{\mathbf{\bar{v}}_0}}
\newcommand{\vibar}{\textcolor{Red}{\mathbf{\bar{v}}_i}}

\renewcommand{\v}{\mathbf{v}}
\newcommand{\vhat}{\mathbf{\hat{v}}}
\newcommand{\vbar}{\mathbf{\bar{v}}}

\newcommand{\score}{\texttt{score}\xspace}
\newcommand{\tol}{\texttt{tol}\xspace}
\newcommand{\abs}{\texttt{abs}\xspace}
\newcommand{\GV}{\texttt{GV}\xspace}

\newcommand{\extract}{\texttt{extract}\xspace}
\newcommand{\iscandidate}{\texttt{is\_candidate}\xspace}

\newcommand{\True}{\textcolor{Purple}{\texttt{True}}\xspace}
\newcommand{\False}{\textcolor{Purple}{\texttt{False}}\xspace}

\newcommand{\res}{\textcolor{Purple}{\texttt{res}}\xspace}

\newcommand{\A}{\textcolor{Blue}{A}\xspace}
\newcommand{\B}{\textcolor{Red}{B}\xspace}
\newcommand{\C}{\textcolor{Green}{C}\xspace}

\newcommand{\Asubscript}{{\textcolor{Blue}{\texttt{A}}}}
\newcommand{\Bsubscript}{{\textcolor{Red}{\texttt{B}}}}

\newcommand{\x}{\ensuremath{\textcolor{Blue}{x}}\xspace}
\newcommand{\y}{\ensuremath{\textcolor{Red}{y}}\xspace}

\newcommand{\ak}{\textcolor{Green}{a_k}}
\newcommand{\bk}{\textcolor{Green}{b_k}}
\newcommand{\ck}{\textcolor{Green}{c_k}}

\newcommand{\akvec}{\textcolor{Green}{\vec{a}_k}}
\newcommand{\bkvec}{\textcolor{Green}{\vec{b}_k}}

\newcommand{\shareA}[1]{\ensuremath{\textcolor{Blue}{[}#1\textcolor{Blue}{]}_\Asubscript}\xspace}
\newcommand{\shareB}[1]{\ensuremath{\textcolor{Red}{[}#1\textcolor{Red}{]}_\Bsubscript}\xspace}

\newcommand{\akA}{\shareA{\ak}}
\newcommand{\bkA}{\shareA{\bk}}
\newcommand{\ckA}{\shareA{\ck}}
\newcommand{\akB}{\shareB{\ak}}
\newcommand{\bkB}{\shareB{\bk}}
\newcommand{\ckB}{\shareB{\ck}}

\newcommand{\resA}{\shareA{\res}}
\newcommand{\resB}{\shareB{\res}}

\newcommand{\xA}{\shareA{\x}}
\newcommand{\xB}{\shareB{\x}}
\newcommand{\yA}{\shareA{\y}}
\newcommand{\yB}{\shareB{\y}}

\newcommand{\MPSPDZ}{\texttt{MP-SPDZ}\xspace}
\newcommand{\mascot}{\texttt{MASCOT}\xspace}
\newcommand{\cowgear}{\texttt{CowGear}\xspace}
\newcommand{\semi}{\texttt{Semi}\xspace}
\newcommand{\hemi}{\texttt{Hemi}\xspace}
\newcommand{\temi}{\texttt{Temi}\xspace}
\newcommand{\soho}{\texttt{Soho}\xspace}
\newcommand{\baseline}{\texttt{Baseline}\xspace}

\newcommand{\sfix}{\texttt{sfix}\xspace}

\newcommand{\secureadd}{\ensuremath{(+)}\xspace}
\newcommand{\securemul}{\ensuremath{(\times)}\xspace}
\newcommand{\securediv}{\ensuremath{(\div)}\xspace}

\newcommand{\Tmeas}{\tilde{\textbf{T}}}
\newcommand{\TmeasOdom}{\textcolor{Blue}{\Tmeas_{\text{odom}}}}
\newcommand{\TmeasLCIntra}{\textcolor{Blue}{\Tmeas_{\text{LC}}}}
\newcommand{\TmeasLCInter}{\textcolor{Purple}{\Tmeas_{\text{LC}}}}

\renewcommand{\v}{\mathbf{v}}

\newcommand{\apriori}{\textit{a priori}\xspace}
\newcommand{\cpp}{\texttt{C++}\xspace}

\newcommand{\Sone}{\texttt{S1}\xspace}
\newcommand{\Stwo}{\texttt{S2}\xspace}
\newcommand{\Sthree}{\texttt{S3}\xspace}
\newcommand{\Sfour}{\texttt{S4}\xspace}
\newcommand{\Sfive}{\texttt{S5}\xspace}
\newcommand{\Ssix}{\texttt{S6}\xspace}
\newcommand{\Sseven}{\texttt{S7}\xspace}
\newcommand{\Seight}{\texttt{S8}\xspace}
\newcommand{\Snine}{\texttt{S9}\xspace}
\newcommand{\Sten}{\texttt{S10}\xspace}

\newcommand{\Done}{\texttt{D1}\xspace}
\newcommand{\Dtwo}{\texttt{D2}\xspace}
\newcommand{\Dthree}{\texttt{D3}\xspace}
\newcommand{\Dfour}{\texttt{D4}\xspace}
\newcommand{\Dfive}{\texttt{D5}\xspace}

\newcommand{\onlinephase}{\textit{online phase}\xspace}
\newcommand{\offlinephase}{\textit{offline phase}\xspace}

\usepackage{caption}
\IEEEoverridecommandlockouts %

\title{\LARGE \bf
CILC: Cryptographically-secure Inter-agent Loop Closure Candidate Detection for Multi-Agent Collaborative SLAM
}

\author{Andrew Fishberg$^{1}$, Yixuan Jia$^{1}$, Jonathan P. How$^{1}$%
\thanks{*~This work is supported by ARL DCIST under Cooperative Agreement Number W911NF-17-2-0181.}%
\thanks{$^{1}$ Department of Aeronautics and Astronautics, Massachusetts Institute of Technology, Cambridge, MA 02139 USA.
        {\tt\small \{fishberg,yixuany,jhow\}@mit.edu}}%
}

\begin{document}

\maketitle
\thispagestyle{empty}
\pagestyle{empty}

\begin{abstract}
Multi-agent Simultaneous Localization and Mapping (SLAM) and collaborative SLAM (CSLAM) require robots to continuously exchange global descriptors (GDs) to detect inter-agent loop closures (ILCs).
While encrypted radios protect this traffic from external eavesdroppers, they offer no protection against a compromised swarm member.
We show this threat is concrete by demonstrating how a corrupted agent can reconstruct approximations of an honest agent's imagery and trajectory from its public GD broadcasts.
To address this, we propose CILC (Cryptographically-secure Inter-agent Loop Closure candidate detection), a first-of-its-kind system leveraging Secure Multi-Party Computation (SMPC) to detect ILC candidates without exchanging GDs in the clear.
Rather than securing the entire CSLAM pipeline, we apply SMPC only to ILC candidate detection (i.e., GD similarity comparison), a privacy-sensitive yet computationally lightweight step, yielding an advantageous privacy-to-overhead trade-off.
We validate in both simulation and hardware experiments that CILC remains real-time and communication-feasible across multimodal GDs (visual and LiDAR), while mitigating information leakage to a compromised swarm agent.
\end{abstract}

\section{Introduction}
\label{sec:introduction}

Multi-agent robotic systems offer significant advantages over single-agent systems, including distributed operation, greater spatial coverage, and improved robustness to individual failures \cite{dorigo2021swarm}. These benefits, however, come at the cost of increased communication and information sharing among agents, creating new privacy and security challenges as larger teams collect, exchange, and coordinate using potentially sensitive data. Recent robot-fleet takeovers and software supply-chain attacks make compromised agents within the swarm a realistic concern for deployed robotic systems \cite{ackerman2025unitree,przymus2025wolves}.

\begin{figure}[t!]
\centering
\includegraphics[width=\linewidth]{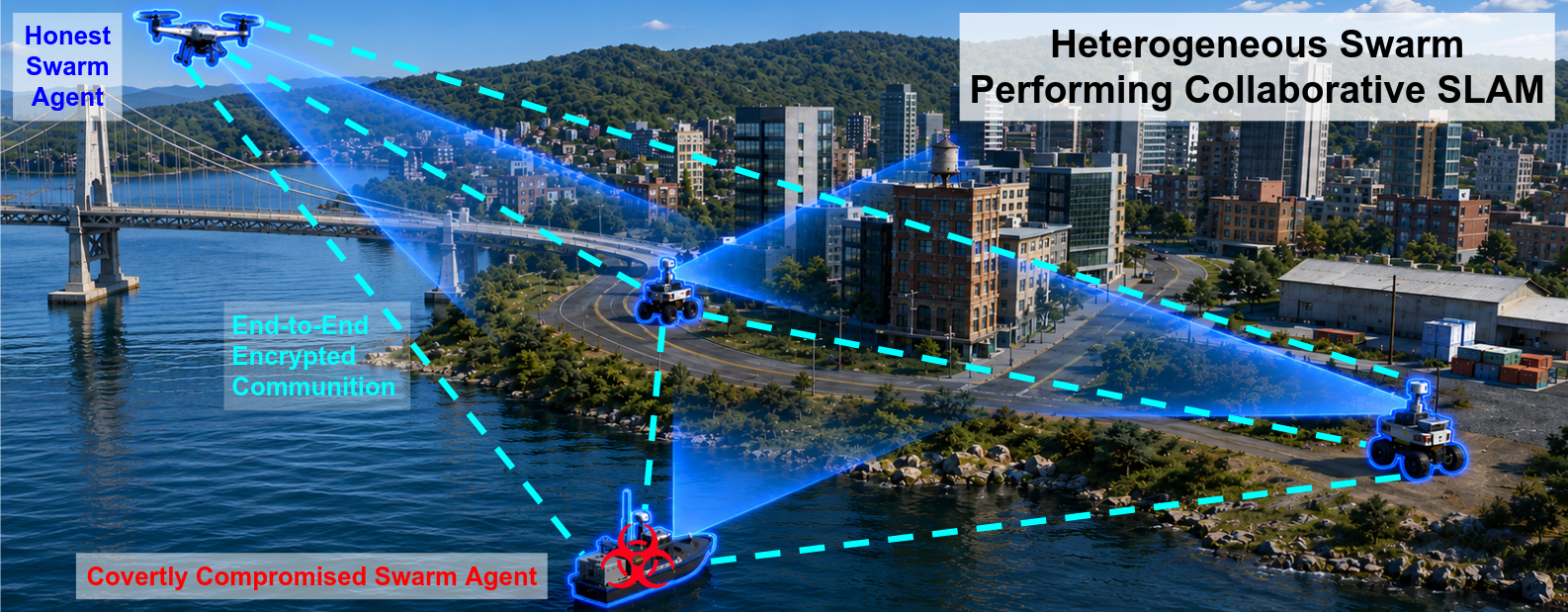}
\caption{Threat model addressed by CILC. A heterogeneous robot team performs \textit{collaborative SLAM}, exchanging observations to detect \textit{inter-agent loop closures} (ILCs). Blue agents are honest collaborators, while the red agent is a compromised but otherwise legitimate swarm member. Transport-layer encryption prevents external eavesdropping, but the compromised agent can access all broadcast data and exfiltrate it to an external adversary without being detected. CILC limits information disclosure by revealing data only when sufficient evidence exists to support an ILC match.}
\label{fig:system-diagram}
\end{figure}

While virtually all modern multi-agent robotic systems rely on encrypted inter-agent communication, this baseline security model exposes a critical gap in the prevailing threat model. As shown in Figure~\ref{fig:system-diagram}, encryption protects data in transit from external adversaries but provides no protection during computation among coordinating agents. Standard practice therefore focuses primarily on external attacks such as eavesdropping, while remaining vulnerable to compromised agents operating within the swarm. Because robots routinely exchange sensitive information, including images, locations, maps, and trajectories, without treating it as private between agents, a single compromised agent could gain access to the collective data of the entire team. Such an adversary may leak mission-critical information, disrupt coordination, manipulate shared state estimates, or otherwise act maliciously from within the system.

This standard security practice in multi-agent robotics creates a inherent tension between \textit{effective collaboration} and \textit{privacy}: collaboration often requires robots to share information, yet increased sharing expands the potential consequences of data leakage, compromise, or misuse. This tension is particularly acute in multi-agent Collaborative Simultaneous Localization and Mapping (CSLAM), where robots must continuously exchange observations to jointly estimate maps and trajectories while maintaining a consistent swarm-wide representation of large-scale, dynamic environments. While unrestricted information sharing among trusted peers is often assumed to be acceptable under the assumption of complete trust, this assumption often breaks down in deployed, real-world systems, particularly as the swarm’s attack surface expands with increasing robot heterogeneity and the number of fielded agents. Without information compartmentalization, the compromise of any single agent, including one that was initially trustworthy, can expose information about the entire swarm.

\begin{figure}[t!]
\centering
\includegraphics[width=\linewidth]{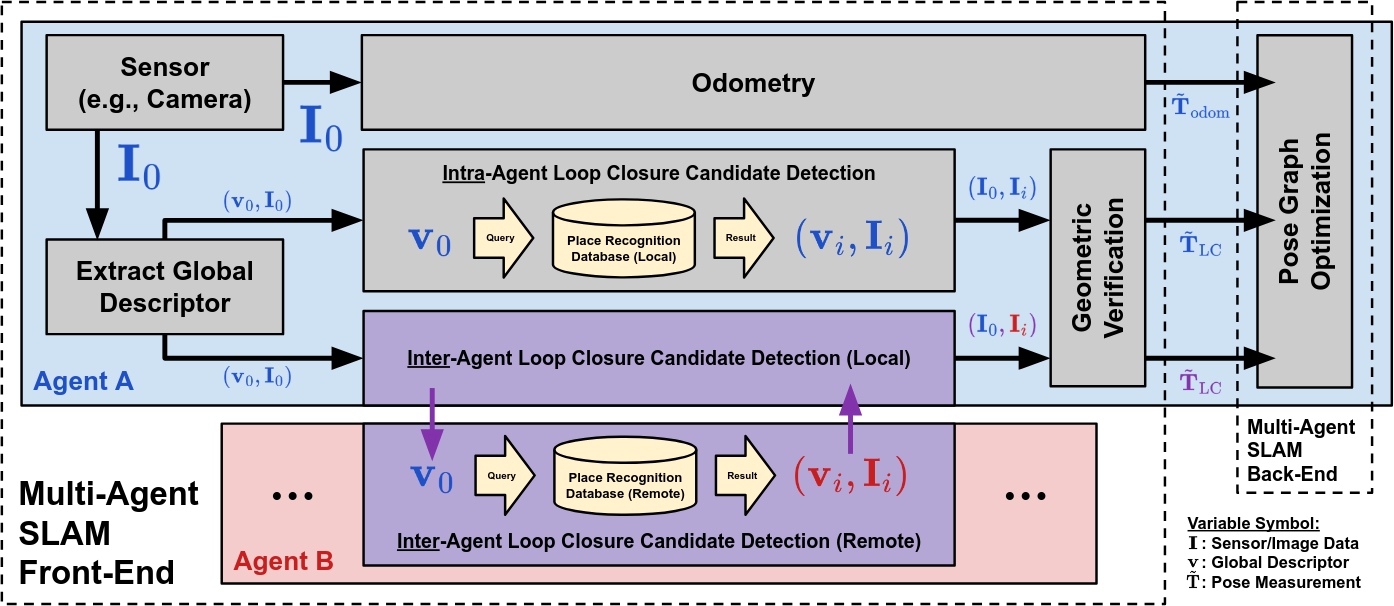}
\caption{A typical CSLAM pipeline between Agents \A and \B, shown from \A's perspective with emphasis on the front-end. The front-end converts raw sensor data $\Iz$ into relative pose measurements $\TmeasOdom, \TmeasLCIntra, \TmeasLCInter \in \SE(3)$, which the back-end consumes for pose-graph optimization. $\TmeasOdom$ and $\TmeasLCIntra$ are computed locally on \A (thus, blue), whereas $\TmeasLCInter$ require data from both agents (thus, purple). $\TmeasLCInter$ are found by broadcasting compact \textit{global descriptors} (GD) $\vz$ across the swarm and cheaply comparing them against $\vi$ to flag \textit{inter-agent loop closure} (ILC) \textit{candidates}; higher-bandwidth data $\Ii$ and expensive \textit{geometric verification} $\GV(\Iz,\Ii)$ follow only when $\score(\vz,\vi) \ge \tol$. CILC (this work), reduces data leakage from the $\vz$ broadcast during ILC candidate detection.}
\label{fig:slam-flowchart}
\end{figure}

Privacy concerns are not new in SLAM research: \cite{choudhary2017distributed,tian2020asynchronous,tian2021distributed,chang2021kimeramulti} explicitly identify privacy as a key motivator for leveraging distributed optimization. However, these works focus primarily on the SLAM back-end (i.e., using distributed optimization to keep inter-agent pose data private during pose-graph optimization) while overlooking the upstream front-end that generates the measurements the back-end consumes. Specifically, \textit{inter-agent loop closures} (ILC) -- i.e., relative pose measurements between agents detected by the CSLAM front-end -- are essential for aligning independently estimated trajectories, but detecting them requires agents to compare observations against one another. In practice, this is done by continuously broadcasting compact \textit{global descriptors} (GD) across the swarm \cite{tian2022kimeramulti,cieslewski2017efficient} to inexpensively flag \textit{ILC candidates}, so that higher-bandwidth data exchange and expensive \textit{geometric verification} (GV) occur only when a candidate pair warrants it (Figure~\ref{fig:slam-flowchart}). This broadcast step, however, can expose observational information that may reveal agent locations, environmental structure, or mission intent to any compromised swarm member (Figure~\ref{fig:system-diagram}). Consequently, protecting GD matching provides a disproportionate privacy benefit while modifying only a small portion of the overall CSLAM pipeline, namely a simple vector-comparison operation.

Recent advances in \textit{Secure Multi-Party Computation} (SMPC)\footnote{We use the term \textit{Secure Multi-Party Computation} (SMPC) over the more common \textit{Multi-Party Computation} (MPC) to emphasize its security and avoid confusion with Model Predictive Control.} offer a promising path forward. As a \textit{security mechanism} developed and popularized by the cryptography community, SMPC enables multiple agents to jointly perform computations while formally limiting what information is revealed to each participant. This in turn supports novel \textit{privacy policies} and threat models that are either overlooked or simply unachievable in multi-agent robotics and CSLAM under encrypted communication alone. In this work, we leverage SMPC for privacy-preserving ILC detection in CSLAM. Specifically, we propose a framework that allows agents to identify ILC candidates without continuous disclosure of observational data, only revealing information when sufficient evidence of a match exists. By reconciling the competing demands of collaboration and privacy, our approach enables efficient and robust cooperation in adversarial multi-agent environments.
This paper makes three key contributions:
\begin{enumerate}[leftmargin=1.5em]
\item We analyze SMPC for CSLAM, identifying ILC candidate detection as a privacy-sensitive yet computationally lightweight stage that is particularly well suited to secure computation (Section~\ref{sec:bridging}).
\item We demonstrate a privacy vulnerability often overlooked in the CSLAM literature (Section~\ref{sec:security-vulnerabilities}).
\item We develop, to our knowledge, the first privacy-preserving ILC candidate detection framework for CSLAM. We validate the proposed approach through simulation and hardware experiments across multiple sensing modalities (Section~\ref{sec:system-architecture-experiments}).%
\end{enumerate}

\section{Related Works}
\label{sec:related-works}

While cybersecurity has historically received limited attention in multi-agent robotics, interest is growing rapidly \cite{bottaCyberSecurityRobots2023}. Privacy has a longer history in controls, particularly through differential privacy \cite{dwork2014algorithmic,leny2014differentially,cortes2016differential,han2017differentially,hanPrivacyControlDynamical2018a,benvenutiGuaranteedFeasibilityDifferentially2024} and homomorphic-encryption-based control and optimization methods \cite{shoukry2016privacyaware}. Although related by ``privacy'' or ``trust’’ concerns, SMPC \cite{evans2018pragmatic}, differential privacy \cite{dwork2014algorithmic}, blockchain \cite{shi2020consensus}, and reputation-based methods \cite{josang2007survey} address distinct orthogonal issues.

SMPC is a mature technology, with theoretical foundations dating to the 1980s \cite{evans2018pragmatic} and practical implementations emerging in the 2010s \cite{hastings2019sok,keller2020mpspdz}. Hastings et al.'s SoK \cite{hastings2019sok} remains a useful overview of SMPC approaches, threat models, and terminology. This work uses the more recent \texttt{MP-SPDZ} \cite{keller2020mpspdz} library, an actively developed SMPC compiler that supports multiple protocols with differing security guarantees and performance characteristics, along with their associated primitives (Section~\ref{sec:bridging}).
SMPC has also been applied in robotics-adjacent settings including cloud-based control, satellite deconfliction, multi-robot task allocation, secure neural-network inference, and secure localization \cite{alexandru2020secure,kammSecureFloatingPoint2015,alsayegh2022PrivacyPreserving,diamond2025optimizing,choncholasSnailSecureSingle2024}.

To the best of our knowledge, this paper is the first to directly integrate SMPC into a multi-agent CSLAM pipeline by applying it to ILC candidate detection, although several robotics and computer vision researchers have studied closely related problems.
For example, Pollefeys et al. have a long line of work on privacy-preserving image queries, localization, and structure-from-motion (SfM), offering a promising non-SMPC path toward private ILC detection \cite{speciale2019privacy,speciale2019privacya,geppert2020privacy,dusmanu2021privacypreserving,geppert2021privacy,geppert2022privacy,pan2023privacy}.
Since conventional image descriptors \cite{galvezlopezDBoW2012,arandjelovic2016netvlad,oquab2024dinov2,simeoni2025dinov3} are not designed with privacy in mind, any data obfuscation they provide is merely a byproduct of being compact (lossy) vectorized image representations. This is not the same as privacy, and it does not prevent information leakage through the GD broadcast step of a multi-agent CSLAM pipeline (Section~\ref{sec:security-vulnerabilities}). By contrast, Pollefeys and colleagues pursue privacy by design, developing image representations that intentionally obfuscate visual information through transformations such as line-cloud descriptors \cite{geppert2021privacy,geppert2020privacy}. However, some of these approaches have subsequently been shown to remain vulnerable to image reconstruction attacks \cite{chelani2021how}. More fundamentally, they require specialized privacy-preserving descriptors rather than leveraging modern state-of-the-art semantic representations, potentially sacrificing place-recognition performance relative to descriptors such as DINOv3 \cite{simeoni2025dinov3}.

Most similar to our approach is Snail \cite{choncholasSnailSecureSingle2024}, a privacy-preserving localization system that uses SMPC to localize a robot without revealing either the query image or map data. In contrast, our work addresses multi-agent CSLAM, where agents jointly compute over private data to build and localize within a shared map. Rather than protecting the entire pipeline, we apply SMPC only to ILC candidate detection. Geometric verification is then performed in the clear only when a privately computed similarity score exceeds a threshold, preventing information leakage when no loop closure exists. Because this stage is computationally lightweight (essentially a vector comparison), it delivers substantial privacy benefits with far less overhead than protecting the full pipeline.

In summary, prior work has explored privacy in the CSLAM back-end, privacy-preserving visual representations, and SMPC for related robotics problems. However, none uses SMPC for privacy-preserving inter-agent loop closure candidate detection in multi-agent CSLAM, which is the central contribution of this paper.

\section{Bridging SMPC \& CSLAM}
\label{sec:bridging}

\begin{figure*}[t!]
\centering
\hfill
\begin{subfigure}{.29\linewidth}
\centering
\includegraphics[height=1.6cm, keepaspectratio]{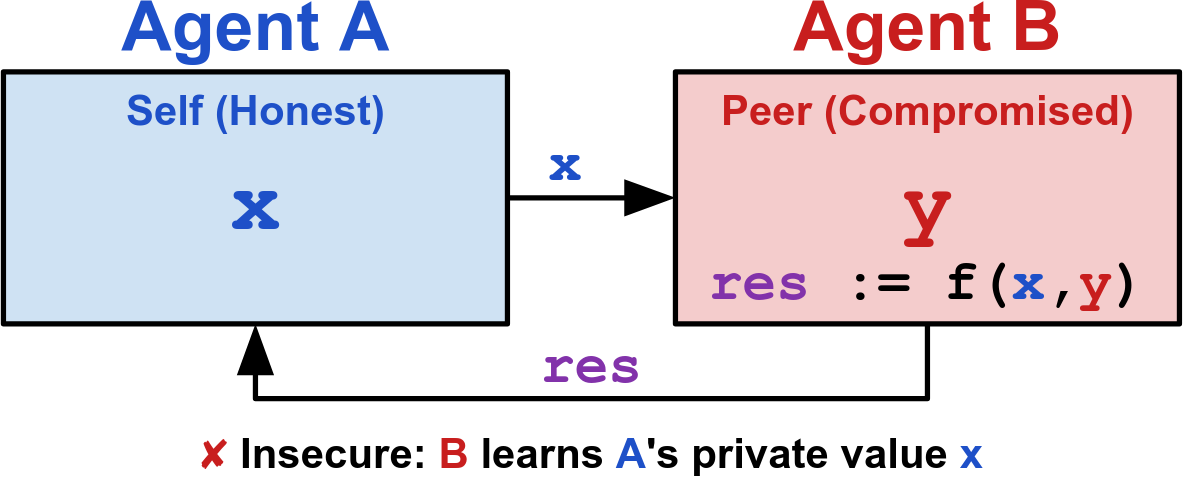}
\vspace*{\fill}
\caption{Full-Trust Paradigm}
\label{fig:mp-insecure}
\end{subfigure}
\hfill
\begin{subfigure}{.42\linewidth}
\centering
\includegraphics[height=1.6cm, keepaspectratio]{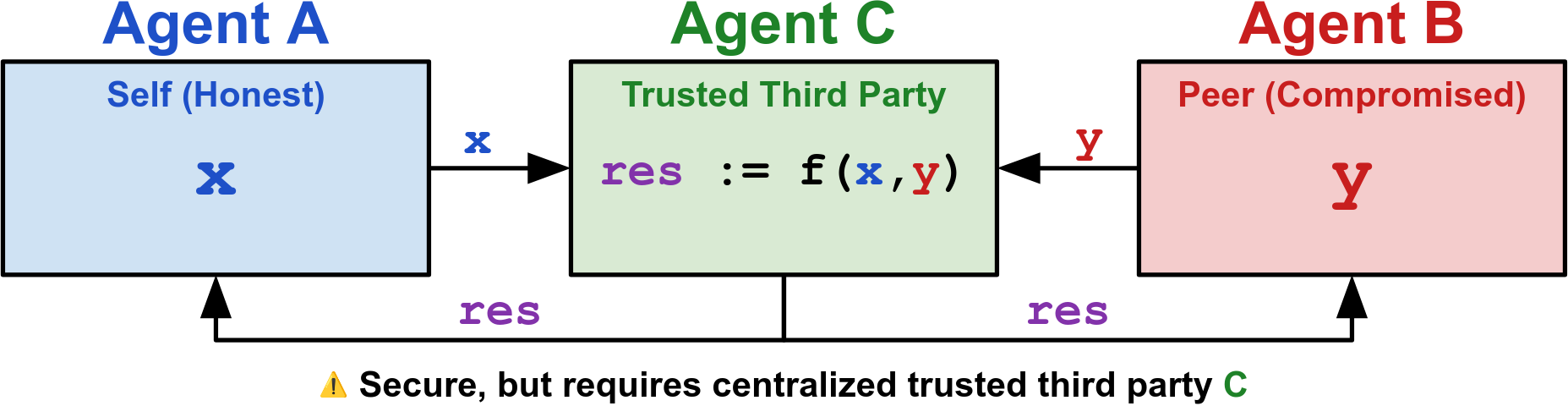}
\vspace*{\fill}
\caption{Trusted Third Party (TTP) Paradigm}
\label{fig:mp-third-party}
\end{subfigure}
\hfill
\begin{subfigure}{.27\linewidth}
\centering
\includegraphics[height=1.6cm, keepaspectratio]{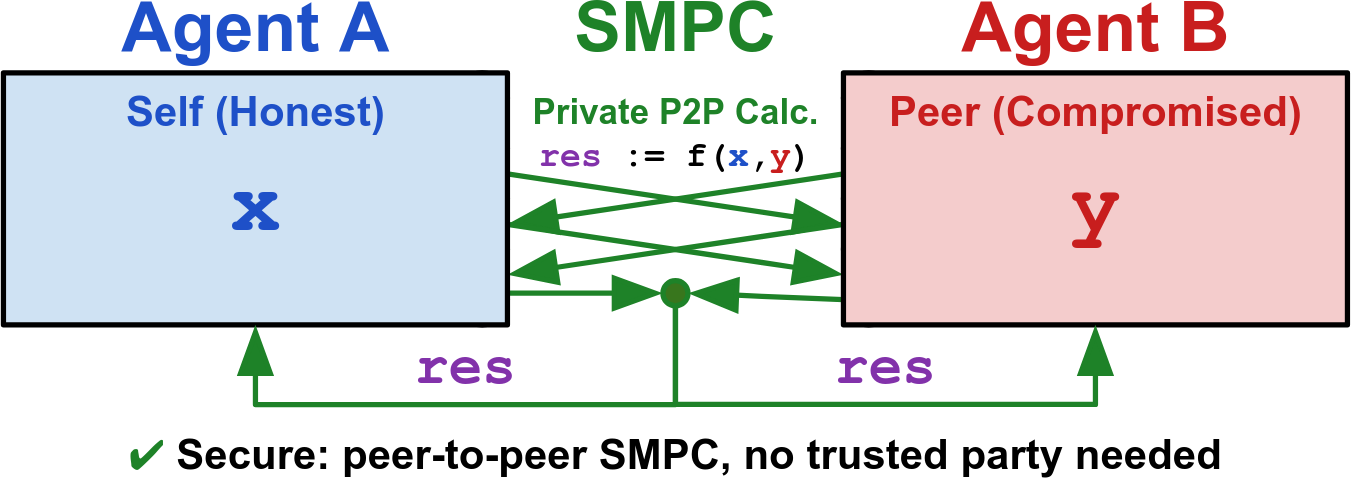}
\vspace*{\fill}
\caption{SMPC Paradigm \textbf{[Our Approach]}}
\label{fig:mp-smpc}
\end{subfigure}
\hfill
\vspace*{\fill}
\caption{
Three paradigms for the joint calculation of $\res := f(\x,\y)$, where \res depends on both \A's private input \x and \B's private input \y. All arrows represent encrypted inter-agent communication, so every approach is protected from external eavesdroppers.
\textbf{(a)}~\A shares \x directly with \B, who computes and returns \res. This approach is simple and computationally efficient, since all computation is done in the clear, but it exposes \x to \B in plaintext and thus requires \A to \textit{fully trust} \B with its private input \x and to honestly compute and return \res.
\textbf{(b)}~\A and \B outsource the computation to a mutually trusted third party (TTP), here \C. Both parties provide their private inputs to \C in plaintext, and \C computes \res in the clear and returns only \res to \A and \B. Since only \C accesses the private inputs, \A cannot learn \y and \B cannot learn \x. When such a trusted entity is available, this approach is simple and efficient. However, establishing trust in a third party can be difficult in the field, and reliance on a secured centralized node is often operationally undesirable or infeasible in practice.
\textbf{(c)}~\A and \B perform SMPC, which provides the same privacy guarantees as a TTP without requiring one (i.e., SMPC can be thought of as creating a ``virtual trusted third party'' through peer-to-peer distributed computation). The tradeoff is that, without an actual TTP, the agreed-upon function cannot be computed in the clear, incurring additional privacy-protecting overhead.
}
\label{fig:multiparty}
\end{figure*}

A full treatment of SLAM \cite{carlone2025slam} and SMPC \cite{evans2018pragmatic,hastings2019sok} is beyond our scope, but a high-level understanding of both motivates our system and its design choices.
This work leverages \texttt{MP-SPDZ} \cite{keller2020mpspdz}, a mature library compiling simple scripts to over 25 SMPC protocols. Throughout, we color-code Agent \A's private values \textcolor{Blue}{blue} (e.g., $\x$), Agent \B's \textcolor{Red}{red}, jointly-computed values \textcolor{Purple}{purple} (e.g., $\res := f(\x,\y)$), and secure values \textcolor{Green}{green} (e.g., Beaver Triples $(\ak, \bk, \ck)$, see Section~\ref{sec:smpc-primitives}). CSLAM ILC candidate detection via GD is discussed in Section~\ref{sec:slam-global-descriptors}.

\subsection{SMPC: Overview, Key Guarantees, \& Limitations}
\label{sec:smpc-overview}

SMPC enables mutually distrusting parties to jointly compute a function over their private inputs, peer-to-peer, without revealing those inputs to one another (Figure~\ref{fig:multiparty}). Within an assumed adversary model (Section~\ref{sec:smpc-adversary-models}), it provides two key guarantees: \textbf{(i)}~privacy of inputs and intermediate values during computation, and \textbf{(ii)}~correctness of the output with respect to the provided inputs \cite{evans2018pragmatic}. Both have important limitations.

Privacy \textbf{(i)} is bounded by what the output reveals: SMPC cannot prevent leakage through the agreed-upon output itself. If Agents \A and \B compute the mean $\res := \frac{\x + \y}{2}$ of their private values, either party can recover the other's input from $\res$ and its own. 

Correctness \textbf{(ii)} holds with respect to supplied inputs, but does not imply input integrity: a party submitting a fabricated input still has the function faithfully computed over it. Enforcing integrity requires an orthogonal mechanism, e.g., a reputation system \cite{josang2007survey}.

\subsection{SMPC: Primitives, Operations, and Overhead}
\label{sec:smpc-primitives}

\begin{table}[t]
\centering
\caption{Cost profile of SMPC primitives in a two-party setting. Communication Rounds and Field Elements Sent refer to \textit{online phase} (i.e., Beaver Triples are generated \textit{a priori} in \textit{offline phase}).}
\resizebox{\linewidth}{!}{\begin{tabular}{lccc}
\hline
\textbf{Operation} &
\makecell{\textbf{Comm.}\\\textbf{Rounds}} &
\makecell{\textbf{Field}\\\textbf{Elements Sent}} &
\makecell{\textbf{BTs}\\\textbf{Consumed}} \\
\hline
Secret Share                   & 1   & 1    & 0 \\
Secure Addition                & 0   & 0    & 0 \\
Secure Multiplication          & 1   & 2    & 1 (scalar) \\
Secure Dot Product (Naive)     & $N$ & $2N$ & N (scalar) \\
Secure Dot Product (Efficient) & 1   & $2N$ & 1 (vector) \\
\hline
\end{tabular}}
\label{tab:smpc-costs}
\end{table}

SMPC is often performed over $\fields_p$, a finite prime field of order $p$, using two primitive operations: \textit{Secure Addition} and \textit{Secure Multiplication}. Any conventionally computable function can be realized under SMPC \cite{evans2018pragmatic}, so the central concern is not expressiveness but computational overhead, since private versions of familiar operations and number representations shift sharply in relative cost (e.g., $\securediv \gg \securemul \gg \secureadd$ and $\texttt{float} \gg \texttt{int} = \texttt{fixed}$).

Agents \A and \B computing $\res := f(\x,\y)$ over private $\x, \y \in \fields_p$ (Figure~\ref{fig:mp-smpc}) have three \textit{online} steps to consider:

\textbf{(i) Distribute Input Secret Shares:}
Each agent splits each private variable into uniform random \textit{secret shares}, one per agent. \A maps $\x \mapsto (\xA, \xB)$, keeping $\xA$ and sending $\xB$ to \B; \B maps $\y \mapsto (\yA, \yB)$ likewise. Shares are sampled uniformly over $\fields_p$, with the owner's final share chosen so all shares sum (mod $p$) to the original value (e.g., $\xA := \x - \xB$ implying $\x = \xA + \xB$). Each share thus reveals nothing on its own. A hidden value can be \textit{opened} at any time by having all agents broadcast and sum over all shares.

\textbf{(ii) Secure Computation:}
Each agent computes its shares of the output (and all intermediate values) using secure operations, which preserves the privacy (i.e., the uniform-distribution property) of shares throughout. As typically implemented \cite{evans2018pragmatic,keller2020mpspdz}, Secure Addition is an entirely local (inexpensive) operation, while Secure Multiplication requires one round of communication between agents per multiplication (expensive). The difficulty of Secure Multiplication arises because the product of two secret-shared values depends on shares held by other agents, so it cannot be computed locally. To resolve this without revealing any agent's shares, a \textit{Beaver Triple} (BT) -- a tuple of secret-shared helper values $(\ak, \bk, \ck)$ s.t. $\ck = \ak\bk$ -- allows the agents to open masked versions of the private terms that can represent the product's value without directly exposing the terms.

Additionally, many structured calculations admit further optimizations. For example, the Private Dot Product of two $N$-vectors (useful in Section~\ref{sec:cilc-protocol-design}) needs a single communication round using one \textit{vector BT} -- $(\akvec, \bkvec, \ck)$ s.t. $\ck = \akvec \cdot \bkvec$ -- instead of $N$ element-wise multiplications. Table~\ref{tab:smpc-costs} shows key primitives and costs.

Because BTs are independent of the parties' inputs, they can be bulk-generated \apriori during an \textit{offline phase}, separate from steps \textbf{(i)}--\textbf{(iii)} (the \textit{online phase}).
This distinction is standard in the SMPC literature and critical for \textit{online} performance \cite{evans2018pragmatic}.
For instance, an \textit{offline-only trusted party} (OOTP) -- an honest server available only before, but not during, deployment -- can trivially generate and distribute BTs ahead of time.\footnote{An OOTP can directly generate 5 uniform random values $\akA, \akB, \bkA, \bkB, \ckA \in \fields_p$, selecting the last s.t. $\ckB := (\akA + \akB)(\bkA + \bkB) - \ckA$, then distribute the shares $(\akA, \bkA, \ckA)$ to \A and $(\akB, \bkB, \ckB)$ to \B accordingly.}
In the absence of an OOTP (or when additional BTs are needed at runtime), parties can still generate BT shares in a secure peer-to-peer manner, although doing so requires expensive cryptographic protocols such as Oblivious Transfer (OT) \cite{evans2018pragmatic}.
In practice, this is orders of magnitude more costly than the entire SMPC \textit{online phase} in terms of both runtime and communication, and is ideally avoided when possible.
Fortunately, in many practical applications (e.g., robotics), a pre-mission OOTP is typical (e.g., to distribute radio encryption keys) and can thus be additionally tasked to efficiently generate BTs during the \textit{offline phase}.

\textbf{(iii) Opening Result:} Once each agent holds its share of the output ($\resA$, $\resB$), the result is publicly opened ($\res := \resA + \resB$). Only the final output is opened, keeping inputs and intermediates hidden.

\subsection{SMPC: Protocols and Threat Models}
\label{sec:smpc-adversary-models}

\begin{table}[t]
  \caption{SMPC adversarial models \cite{evans2018pragmatic} and the corresponding relevant \MPSPDZ protocols implementing each (i.e., over a prime field with a dishonest-majority corruption model) \cite{keller2020mpspdz}. The listed protocols are those our system is benchmarked with in Section~\ref{sec:system-architecture-experiments}.}
  \label{tab:smpc-adversary-models}
  \centering
  \resizebox{\linewidth}{!}{
  \begin{tabular}{@{}c@{\hspace{1em}} l p{7cm} l}
    \toprule
    & \textbf{Model} & \textbf{Adversary Model Description} & \textbf{Protocol(s)} \\
    \midrule
    \addlinespace[0.4em]
    \multirow[c]{3}{*}{\rotatebox[origin=c]{90}{\parbox{2.25cm}{\centering More Secure \rightarrowfill \\ \leftarrowfill Less Overhead}}} &
    \makecell[tl]{Malicious \\[-3pt] \textit{\textcolor{gray}{a.k.a. ``active''}}} & Corrupted parties may deviate arbitrarily, while honest parties always get correct output or abort. & \mascot \\
    \addlinespace[0.4em]
    & Covert & Corrupted parties may deviate arbitrarily, but are caught with fixed probability, a probabilistic deterrent. & \cowgear \\
    \addlinespace[0.4em]
    & \makecell[tl]{Semi-Honest \\[-3pt] \textit{\textcolor{gray}{a.k.a. ``passive''  -or-}} \\[-3pt] \textit{\textcolor{gray}{``honest-but-curious''}}} & Corrupted parties follow protocol honestly, but cannot learn more than the output. & \makecell[tl]{\semi, \hemi,\\ \temi, \soho} \\
    \addlinespace[0.4em]
    \bottomrule
  \end{tabular}}
\end{table}

\MPSPDZ compiles to over 25 SMPC protocols, each named and categorized by the field it operates over and its threat model. The \textit{adversary model} specifies \textit{how agents may misbehave} (Table~\ref{tab:smpc-adversary-models}), while the \textit{corruption model} specifies \textit{how many agents are corrupted} ($\geq$50\% \textit{dishonest majority} vs. $<$50\% \textit{honest majority}). Lower security requirements permit more efficient protocols. Our application uses prime fields and assumes a \textit{dishonest majority} (\textit{honest majority} is moot with two parties). Table~\ref{tab:smpc-adversary-models} lists the most relevant protocols (which we benchmark our system with in Section~\ref{sec:system-architecture-experiments}) and their respective adversary-model assumptions.

\subsection{SLAM: Global Descriptor (GD) Similarity Scoring}
\label{sec:slam-global-descriptors}

Figure~\ref{fig:slam-flowchart} depicts a typical CSLAM pipeline. From Agent \A's perspective, ILC step has a raw observation $\Iz$ (e.g., image, LiDAR scan) processed to \textit{extract} a compact GD vector:
$\vz := \extract(\Iz)$.
GDs are broadcast across the swarm, where they are \textit{scored} for similarity $\score(\vz,\vi)$ against other agents' GDs $\vi$. An ILC candidate is flagged when GD similarity exceeds $\tol$:
$$\iscandidate(\vz,\vi) := \score(\vz,\vi) \ge \tol$$
Only then do higher-bandwidth data $\Ii$ and expensive \textit{geometric verification} $\GV(\Iz,\Ii) \mapsto \TmeasLCInter$ follow. Each type of GD defines its own $\extract$ and $\score$.

We consider the following image-based GDs: DBoW2 \cite{galvezlopezDBoW2012} (a ``classical'' bag-of-words descriptor popularized by ORB-SLAM \cite{murartalORBSLAM2015}), NetVLAD \cite{arandjelovic2016netvlad} (a learned CNN descriptor bridging classical and transformer-based approaches), DINOv2 \cite{oquab2024dinov2} (a self-supervised vision transformer foundation model), and DINOv3 \cite{simeoni2025dinov3} (a more recent, state-of-the-art version). To demonstrate multimodality, we also consider Scan Context \cite{kim2018scan} (a LiDAR-based GD): our implementation (Section~\ref{sec:system-architecture-experiments}) is agnostic to the sensor type and $\extract$ function, only needing to implement $\iscandidate$ (i.e., $\score$) over $\vz$ and $\vi$. Table~\ref{tab:global-descriptors} details each considered GD and its $\score$.

\begin{table*}[t!]
  \centering
  \caption{Comparison of \textit{global descriptors} (GD) used for visual and LiDAR-based \textit{inter-agent loop closure} (ILC) candidate detection. Vector lengths refer to typical / default configurations as reported in the original works (many expose parameters that change the size).}
  \label{tab:global-descriptors}
  \small
  \setlength{\tabcolsep}{4pt}
  \renewcommand{\arraystretch}{1.2}
  \resizebox{\linewidth}{!}{%
  \begin{tabular}{@{}l l l l l@{}}
    \toprule
    \textbf{ID} &
    \textbf{Descriptor} &
    \textbf{Modality} &
    \textbf{Typical descriptor dimensionality} &
    \textbf{Similarity Score description} \\
    \midrule
    \Done &
    DBoW2 (2012) \cite{galvezlopezDBoW2012} &
    Image &
    Size of vocabulary (e.g., between $10^3$ and $10^6$ words) &
    $L_1$-dist on TF-IDF as $L_1$-unit-vec \\
    \Dtwo &
    NetVLAD (2016) \cite{arandjelovic2016netvlad} &
    Image &
    16k-D or 32k-D, reduced to 4096-D by PCA, whitening, \& normalization &
    Cosine similarity \\
    \Dthree &
    DINOv2 (2023) \cite{oquab2024dinov2} &
    Image &
    2 tokens $\times$ 384-D (S) / 768-D (B) / 1024-D (L) / 1536-D (g) &
    Cosine similarity \\
    \Dfour &
    DINOv3 (2025) \cite{simeoni2025dinov3} &
    Image &
    2 tokens $\times$ 384-D (S/S+) / 768-D (B) / 1024-D (L) / 1280-D (H+) / 4096-D (7B) &
    Cosine similarity \\
    \Dfive &
    Scan Context (2018) \cite{kim2018scan} &
    LiDAR &
    $20 \times 60 = $ 1200-D matrix (radial-by-azimuth bins); 20-D vector (retrieval) &
    Cosine similarity; $L_1$ (retrieval) \\
    \bottomrule
  \end{tabular}
  }
\end{table*}

\section{Demonstration of Security Vulnerabilities}
\label{sec:security-vulnerabilities}

To motivate our security concerns, consider two attacks a compromised swarm agent \B can mount against an honest \A during typical CSLAM operation. As a trusted peer, \B sits on the encrypted communication network, so transport-layer encryption offers \A no protection. Both attacks are undetectable by \A, as they require only listening to \A's ILC GD broadcast (something an honest \B would do anyway) and withholding ILC candidates that only \B knows about.

\subsection{Image Reconstruction from GD Broadcast}
\label{sec:image-reconstruction}

{
\newlength{\figlabelheight}
\setlength{\figlabelheight}{.75em} %

\begin{figure}[t!]
\centering
\begin{tikzpicture}
\node[anchor=south west, inner sep=0] (img) at (0,0)
    {\includegraphics[width=.9\linewidth]{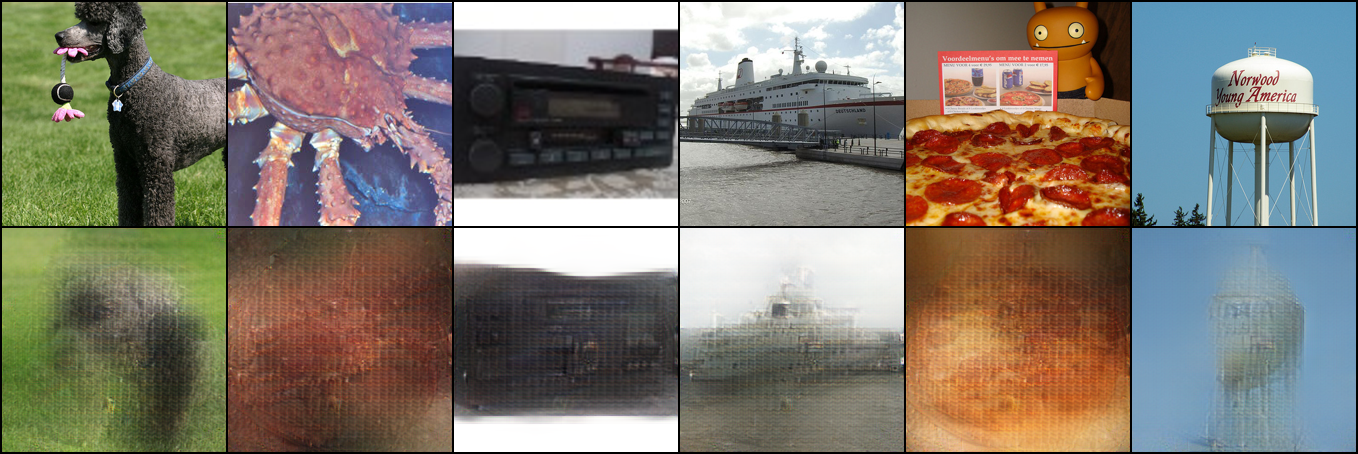}};
\begin{scope}[x={(img.south east)}, y={(img.north west)}]
    \node[rotate=90, anchor=south] at (0.0,0.75)
        {\resizebox{!}{\figlabelheight}{\shortstack{\A's original \\ $\Iz$}}};
    \node[rotate=90, anchor=south] at (0.0,0.25)
        {\resizebox{!}{\figlabelheight}{\shortstack{\B's malicious \\ $\extract^{-1}(\vz)$}}};
\end{scope}
\end{tikzpicture}
\caption{
We demonstrate a simple \textit{image reconstruction attack} on a 1536-D DINOv2 GD. Typical CSLAM pipelines (Figure~\ref{fig:comms-insecure}) are vulnerable to this attack, but CILC (our approach, Figure~\ref{fig:comms-smpc}) protects against this.
\textbf{Top row:}~\A's original ImageNet \cite{deng2009imagenet} images (not seen during training).
\textbf{Bottom row:}~\B's malicious reconstructions of each $\Iz$ from broadcasted $\vz$. Additional details in Section~\ref{sec:image-reconstruction}.}
\label{fig:image-reconstruction}
\end{figure}
}

Figure~\ref{fig:image-reconstruction} demonstrates a basic \textit{image reconstruction attack} on a 1536-D GD made using DINOv2 \cite{oquab2024dinov2}. Merely by listening to Agent \A's GD broadcast, a corrupted Agent \B can unknowingly reconstruct an approximate, lossy version of \A's original image. Each GD concatenates DINOv2 ViT-B/14's classification token (768-D) and the mean of its patch tokens (768-D) into a fixed-length 1536-D vector. We train a convolutional decoder to approximate the inverse mapping $\extract^{-1}$ from descriptor space $\vz$ to image space $\Iz$: a fully-connected layer projects the descriptor into a $512$-channel $7\times7$ feature map, which five upsampling stages (each a residual block, $2\times$ nearest-neighbor upsample, and $3\times3$ convolution) enlarge to a $224\times224$ RGB image. We train for 40 epochs (batch size 16) on the first 114,592 ImageNet training images \cite{deng2009imagenet} against a weighted sum of L1 pixel, LPIPS perceptual \cite{zhang2018unreasonable}, and descriptor-consistency losses, using a single NVIDIA GeForce RTX 2080.
Although this simple decoder may not generalize beyond ImageNet,
it shows that GDs divulge significant information with little effort. This underscores the severity of the vulnerability in CSLAM systems, where GDs are broadcast and not all agents can be trusted.

\subsection{Path Reconstruction from GD Broadcast}
\label{sec:path-reconstruction}

\begin{figure}[b!]
\centering
\begin{minipage}[c]{0.19\linewidth}
    \centering
    \begin{subfigure}{\linewidth}
        \centering
        \includegraphics[width=.85\linewidth]{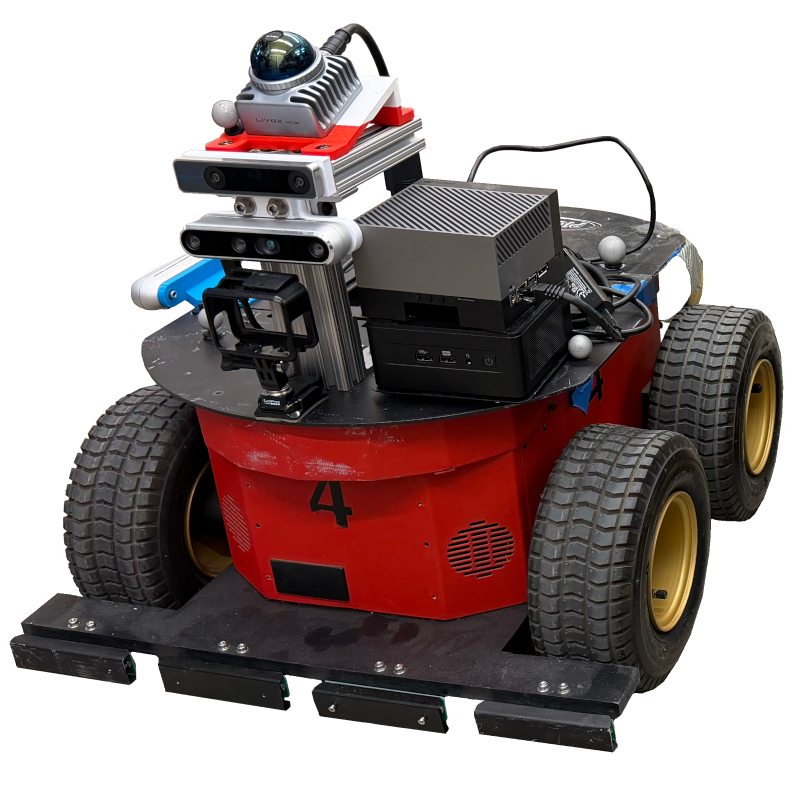}
        \caption{Agent \A}
        \label{fig:agent-a}
    \end{subfigure}

    \vspace{0.5em}

    \begin{subfigure}{\linewidth}
        \centering
        \includegraphics[width=.85\linewidth]{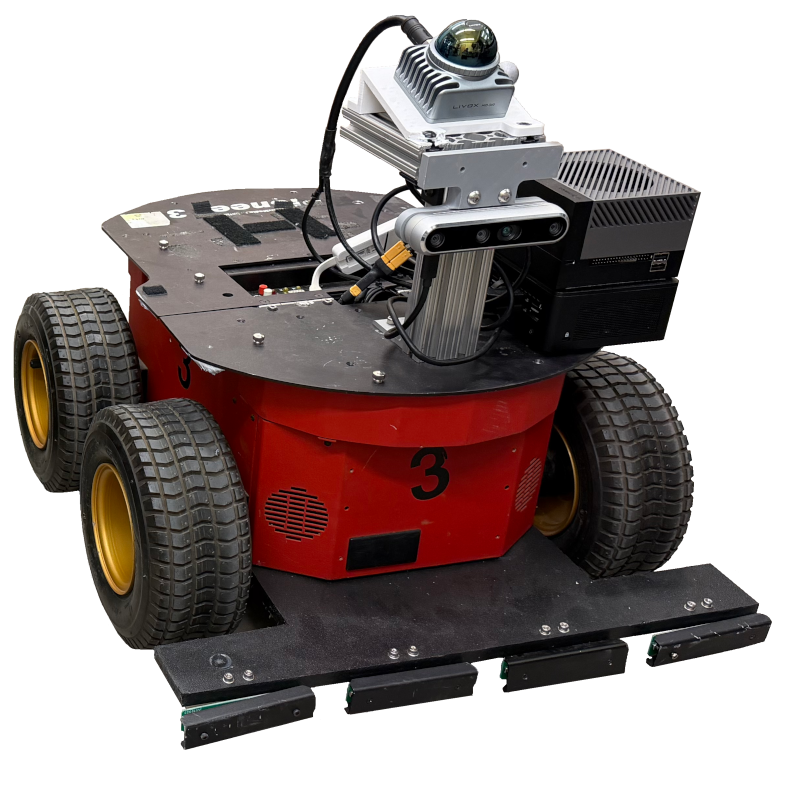}
        \caption{Agent \B}
        \label{fig:agent-b}
    \end{subfigure}
\end{minipage}%
\hfill
\begin{subfigure}[c]{.8\linewidth}
    \centering
    \includegraphics[width=\linewidth]{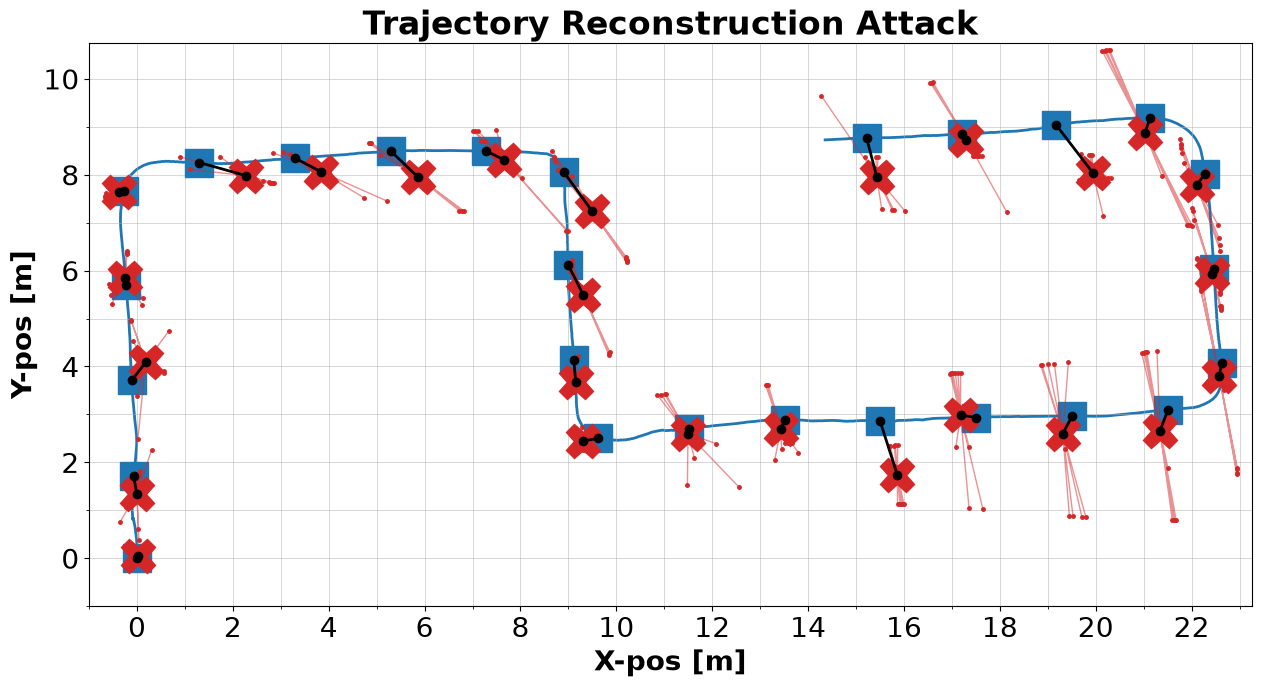}
    \vspace{-1.75em}
    \caption{\B's malicious reconstruction of \A's trajectory}
    \label{fig:reconstruction-attack}
\end{subfigure}

\caption{We demonstrate a simple \textit{trajectory reconstruction attack} using the pictured robot hardware and 1536-D DINOv2 GDs.
\A and \B patrol the same office hallways %
without following the same path.
Nevertheless, \B reconstructs a close approximation of \A's trajectory with minimal effort and without \GV.
Blue Squares show \A's true position when broadcasting a GD $\vz$; Red Dots show positions where \B collected a $\vi$ similar to $\vz$ (i.e., where $\iscandidate(\vz,\vi)$ is $\True$); and Red Xs show \B's malicious reconstructions of \A's positions, computed solely from \B's local knowledge and \A's broadcast GDs by simply averaging the Red Dots. Additional details in Section~\ref{sec:path-reconstruction}.}
\label{fig:path-reconstruction}
\end{figure}

Figure~\ref{fig:path-reconstruction} demonstrates a basic \textit{trajectory reconstruction attack} on real robot hardware using the same DINOv2 GD.
Each agent is a Red Rover robot (Figures~\ref{fig:agent-a} and \ref{fig:agent-b}) equipped with a Livox Mid-360 LiDAR for odometry, a RealSense D455 for ILC detection, and a NUC13 for onboard compute (13th Gen Intel i7-1360P (16) @ 5\;GHz, 64\;GB RAM).
\A and \B patrol the same office environment, while compromised \B maliciously reports $\iscandidate$ as $\False$, preventing \A from receiving any ILCs.
Although \B cannot perform \GV without $\Iz$, it still receives every $\vz$ in the clear.
By comparing each $\vz$ against its own $\vi$ and recording where $\iscandidate$ is $\True$ in its local map, \B can gradually reconstruct \A's approximate trajectory using only \A's public GDs.
This further demonstrates how readily GDs leak sensitive information, reinforcing the findings of Section~\ref{sec:image-reconstruction}.

\begin{figure}[t!]
\centering
\hfill
\begin{subfigure}{\linewidth}
\includegraphics[width=\linewidth]{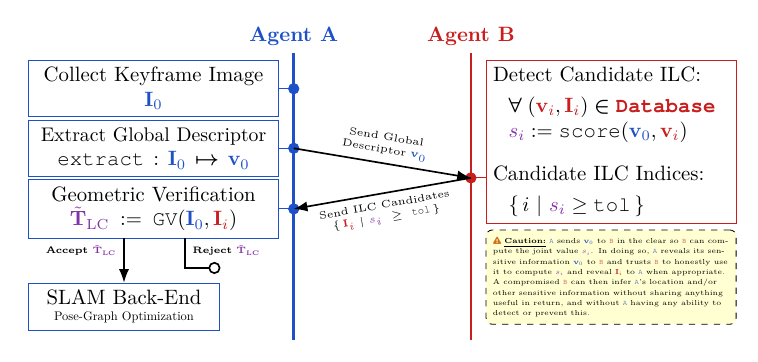} \vspace*{-0.3in}
\caption{Data-leaking ILC protocol \textbf{[Typical Approach]}}
\label{fig:comms-insecure}
\end{subfigure}
\hfill\null

\vspace*{-0.1in}

\hfill
\begin{subfigure}{\linewidth}
\includegraphics[width=\linewidth]{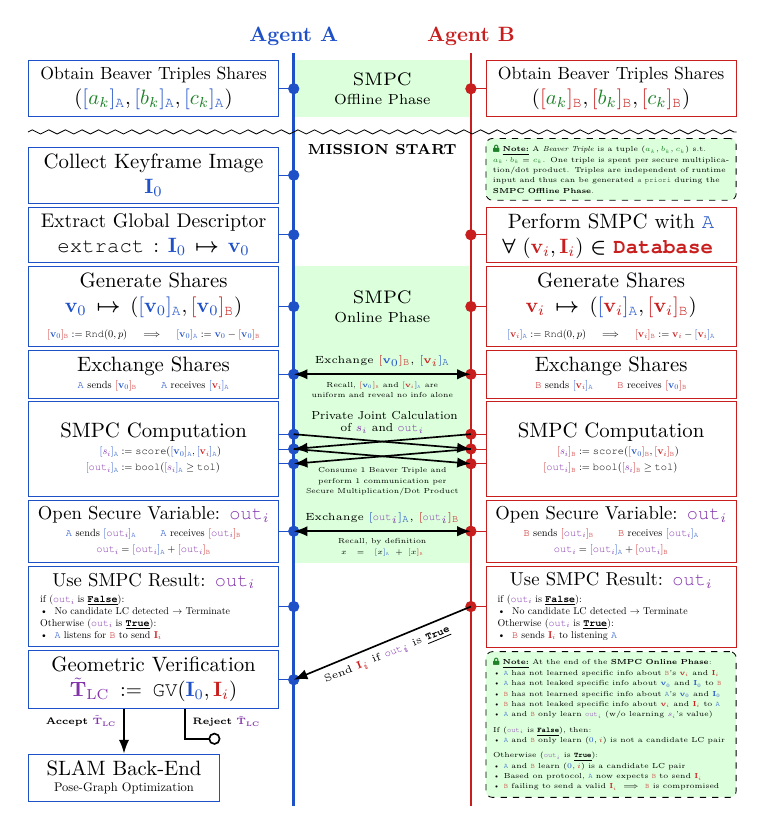} \vspace*{-0.3in}
\caption{Privacy-preserving ILC protocol \textbf{[Our Approach: CILC]}}
\label{fig:comms-smpc}
\end{subfigure}
\hfill\null
\vspace*{-0.1in}
\caption{Comparison between conventional CSLAM ILC candidate detection pipeline and our approach, \textit{CILC}. Details in Section~\ref{sec:cilc-protocol-design}}
\label{fig:comms}
\end{figure}

\section{System Architecture \& Experiments}
\label{sec:system-architecture-experiments}

\subsection{CILC Protocol Design}
\label{sec:cilc-protocol-design}

To address the CSLAM security vulnerabilities of Section~\ref{sec:security-vulnerabilities}, we propose CILC (Cryptographically-secure Inter-agent Loop Closure candidate detection), which uses SMPC to detect ILC candidates without continuous GD disclosure. This mitigates data exfiltration by compromised agents while preserving the benefits of CSLAM.

Figure~\ref{fig:comms} compares a conventional CSLAM ILC pipeline (Figure~\ref{fig:comms-insecure}) with CILC (Figure~\ref{fig:comms-smpc}). Running the entire pipeline under SMPC would offer maximal protection but incur prohibitive overhead (e.g., runtime, and comms usage). Instead, our analysis (Section~\ref{sec:bridging}) identifies ILC candidate detection as a privacy-sensitive yet computationally simple step (\iscandidate is typically a vector-norm threshold check). This makes it well-suited for efficient SMPC, yielding an advantageous privacy to overhead trade-off.

Thus, CILC computes $\iscandidate(\vz,\vi)$ under SMPC, keeping $\vz$, $\vi$, and all intermediate values hidden while revealing only whether the GD similarity threshold was met (\True/\False). \GV still runs in the clear, but the additional information $\Ii$ is exposed only on a candidate match, significantly constraining what a compromised \B can exfiltrate from honest \A.

Under CILC, \B can no longer passively observe $\vz$, learn the exact value of $\score(\vz,\vi)$, or lie about $\iscandidate(\vz,\vi)$: it cannot learn from \A's $\vz$ without supplying its own $\vi$, which in turn reveals information about \B. \B can still supply arbitrary GDs $\vi$ or images $\Ii$\footnote{Recall from Section~\ref{sec:smpc-overview} that SMPC guarantees output correctness with respect to the provided inputs, but cannot compel a compromised party to supply truthful inputs.} to probe what \A has seen, but a successful arbitrary probe requires guessing a high-dimensional (e.g., $8192$-D) vector $\vi$ within \tol of private input $\vz$ -- generally improbable and impractical. Moreover, if $\iscandidate(\vz,\vi)$ returns \True, both agents learn this under SMPC, so \A then expects \B to supply image $\Ii$ as the CILC protocol requires; failure to do so exposes \B as compromised. Although \B cannot be compelled to provide a true $\Ii$ for the supplied $\vi$, evading detection requires supplying an $\Ii$ that is not obviously tampered with and whose corresponding GD falls within \tol of $\vz$ (without knowing $\vz$): once \B reveals $\Ii$ in the clear, \A simply checks that $\Ii\mapsto\vi$ is consistent with $\score(\vz,\vi)\ge\tol$, and any inconsistency immediately identifies \B as compromised. This strongly incentivizes \B to supply only honest GD-image pairs it has actually observed, or risk detection. %

\subsection{Experimental Design}
\label{sec:experimental-design}

We implement CILC in \MPSPDZ (\texttt{v0.4.2}) and evaluate it through benchmarking and hardware experiments. Benchmarking consists of a parameter sweep (10 trials/config) across 7 protocols (the 6 SMPC protocols in Table~\ref{tab:smpc-adversary-models} plus the insecure \baseline), 9 vector sizes ($16 \leq N \leq 8192$), and 10 \score functions (\Sone-\Sten, see Table~\ref{tab:score-functions}), for a total of 6300 trials.
We benchmark \score directly because \iscandidate simply applies a threshold check around the selected \score implementation.
As is typical, BTs are assumed to be generated \apriori during the \offlinephase (Section~\ref{sec:smpc-primitives}). Table~\ref{tab:results} summarizes results most relevant to our discussion\footnote{Results for \Sten are shown for all $N$ because it is the preferred SMPC implementation (Section~\ref{sec:discussion}). Results for all \score functions are shown at $N=16$ to highlight which are impractical under SMPC, even with small $N$. Additional results are shown for $N=1536$ (GD size used in Section~\ref{sec:security-vulnerabilities}) and $N=8192$ (largest considered GD, formed by concatenating DINOv3 7B's 4096-D classification and 4096-D mean patch tokens, similar to how the 1536-D DINOv2 GD was constructed in Section~\ref{sec:image-reconstruction}).} (raw metrics for all trials are available in our code repository). SMPC runtime and communication are measured directly by \MPSPDZ. Secure implementations use \sfix (\MPSPDZ's secure fixed-point type) configured as 64-bit with 16-bit fractional precision. The insecure \baseline is implemented in \cpp. All benchmarks were run on an Intel i9-10900K (20) @ 5.3\,GHz CPU. Hardware experiments repeat the path reconstruction demonstration (Section~\ref{sec:path-reconstruction}, Figure~\ref{fig:path-reconstruction}) using CILC.

{\everymath{\displaystyle}

\newcommand{\purposewidth}{12cm} %
\newcommand{\colsep}{0pt}        %
\newcommand{\eqscale}{.75}      %
\newcommand{\scoreeq}[1]{\scalebox{\eqscale}{$#1$}}

\begin{table}[t]
\caption{Information about the tested $\score(\vz,\vi)$ functions with a brief rationale for inclusion and expected performance under SMPC. The shorthand $\vzbar, \vibar$ and $\vzhat, \vihat$ represent computing the $L_1$- and $L_2$-normalization of $\vz, \vi$ respectively -- i.e., computing $\textstyle \vbar := \v/\|\v\|_1$ and $\textstyle \vhat := \v/\|\v\|_2$ under SMPC for private input $\v$.}
\label{tab:score-functions}
\centering
\setlength{\tabcolsep}{\colsep}
\resizebox{1\columnwidth}{!}{%
\begin{tabular}{@{}m{0.8cm}m{2.5cm}m{2.75cm}m{\purposewidth}@{}}
\toprule
ID & Score Name & $\score(\vz, \vi) :=$ & Purpose \& Analysis \\
\midrule
\Sone &
\texttt{L1\_dist\_bar} &
\scoreeq{\left\| \vzbar - \vibar \right\|_1} &
\makecell[cl]{\Done's default. Under SMPC, expected to be inefficient since each $L_1$ (three total) requires $N$ \\ conditionals (element-wise \abs) and normalizing vecs requires $N$ multiplications each.} \\
\midrule
\Stwo &
\texttt{L1\_dist} &
\scoreeq{\left\| \vz - \vi \right\|_1} &
\makecell[cl]{\Dfive's retrieval default. Equiv to \Sone if $\vz,\vi$ are unit inputs. Under SMPC, expected to be \\ more efficient than \Sone (no normalization), but $L_1$ still requires $N$ conditionals.} \\
\midrule
\Sthree &
\texttt{L2\_dist\_hat} &
\scoreeq{\sqrt{\left(\vzhat - \vihat\right) \cdot \left(\vzhat - \vihat\right)}} &
\makecell[cl]{Common vec similarity. Under SMPC, expected to be inefficient since vec normalization \\ and sqrt. Expected to outperform \Sone's element-wise \abs since $\cdot$ is efficient under SMPC.} \\
\midrule
\Sfour &
\texttt{L2\_dist} &
\scoreeq{\sqrt{(\vz - \vi) \cdot (\vz - \vi)}} &
\makecell[cl]{Common vec similarity. Equiv to \Sthree if $\vz,\vi$ are unit inputs. Under SMPC, expected to be \\ more efficient than \Sthree (no normalization), but still inefficient because sqrt.} \\
\midrule
\Sfive &
\texttt{L2sq\_dist\_hat} &
\scoreeq{\left(\vzhat - \vihat\right) \cdot \left(\vzhat - \vihat\right)} &
\makecell[cl]{Common vec similarity. Under SMPC, expected to be more efficient than \Sthree (no sqrt), but \\ still inefficient because normalizing vecs requires $N$ multiplications each.} \\
\midrule
\Ssix &
\texttt{L2sq\_dist} &
\scoreeq{\left(\vz - \vi\right) \cdot \left(\vz - \vi\right)} &
\makecell[cl]{Common vec similarity. Under SMPC, expected to be extremely efficient since addition \\ does not communicate, $\cdot$ is efficient under SMPC (same overhead as 1 mult), and no sqrt.} \\
\midrule
\Sseven &
\texttt{cossim\_hat} &
\scoreeq{\vzhat \cdot \vihat} &
\makecell[cl]{\Dtwo--\Dfive's default. \Sseven--\Snine are all algebraically equiv but, under SMPC, this is expected to be \\ the least performant of the three due to the ordering of the secure operations.} \\
\midrule
\Seight &
\texttt{cossim\_2sqrt} &
\scoreeq{\frac{\vz \cdot \vi}{\sqrt{(\vz \cdot \vz)} \sqrt{(\vi \cdot \vi)}}} &
\makecell[cl]{\Dtwo--\Dfive's default. \Sseven--\Snine are all algebraically equiv but, under SMPC, this is expected to be \\ the middle-performing due to the ordering of the secure operations but having two sqrts.} \\
\midrule
\Snine &
\texttt{cossim\_1sqrt} &
\scoreeq{\frac{\vz \cdot \vi}{\sqrt{(\vz \cdot \vz)(\vi \cdot \vi)}}} &
\makecell[cl]{\Dtwo--\Dfive's default. \Sseven--\Snine are all algebraically equiv but, under SMPC, this is expected to be \\ the most performant due to the ordering of the secure operations and only having one sqrt.} \\
\midrule
\Sten &
\texttt{dotprod} &
\scoreeq{\vz \cdot \vi} &
\makecell[cl]{Equiv to \Sseven--\Snine if $\vz,\vi$ are unit inputs. Under SMPC, expected to be most-performant of \\ all \Sone--\Sten since only one efficient $\cdot$ is needed (extremely efficient \Ssix still has $N$ additions).} \\
\bottomrule
\end{tabular}
}
\end{table}
}

\subsection{Discussion of Results}
\label{sec:discussion}

First, when \score functions (\Sone-\Sten, see Table~\ref{tab:score-functions}) are compared with a fixed $N$ and secure protocol (see Table~\ref{tab:results}), we see large performance differences, reflecting how familiar mathematical operations have very different relative costs under SMPC.\footnote{Recall from Section~\ref{sec:smpc-primitives} that Secure Addition requires no communication, Secure Multiplication requires one communication round, and a Secure Dot Product requires only one communication round (same as a single Secure Multiplication) regardless of vector length, though it sends $2N$ field elements rather than $2$.} In particular, any \score function that performs vector normalization under SMPC is needlessly expensive because: \textbf{(i)}~normalization requires $N$ element-wise Secure Multiplications after computing magnitude, and \textbf{(ii)}~since normalization depends only on each party's private input, it can instead be performed locally before entering the SMPC calculation. Consequently, \Sone, \Sthree, \Sfive, \Sseven are impractical. Similarly, Cosine Similarity \Sseven--\Snine reduces to a single Dot Product (\Sten) after local $L_2$-normalization, making \Sseven-\Snine impractical. Thus, \Sten is the preferred implementation of Cosine Similarity under SMPC.

Second, even among the remaining candidates (\Stwo, \Sfour, \Ssix, \Sten), performance varies substantially. The $L_1$-distance (\Stwo) is expensive because it requires $N$ Secure Conditionals (i.e., element-wise \abs), while the $L_2$-distance (\Sfour) requires an expensive Secure Square Root. In contrast, $L_2$-distance squared (\Ssix) requires only inexpensive Secure Additions plus an efficient Secure Dot Product, and \Sten is just a Secure Dot Product, making them substantially more efficient under SMPC. Since cosine similarity is the dominant similarity metric among the surveyed global descriptors (Table~\ref{tab:global-descriptors}), \Sten emerges as the preferred SMPC implementation, offering the best combination of practicality and performance.

Finally, comparing \mascot and \hemi demonstrates the expected security-performance trade-off. For \Sten with $N=8192$ (the largest considered GD), \hemi is $2.06\times$ faster and requires roughly one-third less communication than \mascot; whether this security-performance trade-off is acceptable depends on the application and threat model. Compared with the insecure \baseline, \mascot incurs the expected overhead (e.g., $225.63\times$ runtime and $6.01\times$ comms usage for \Sten with $N=8192$). However, because CILC restricts SMPC to ILC candidate detection, the absolute costs remain modest within the overall CSLAM pipeline: \textbf{CILC's \textit{online phase} remains \textit{real-time} and \textit{comms-feasible}} (e.g., 2.70\;ms and 0.075\;MB for securely comparing two of the 1536-D DINOv2-based GDs used in Section~\ref{sec:security-vulnerabilities}, each of which is 0.012\;MB in the clear), leaving other processes, such as \textit{geometric verification} and \textit{pose-graph optimization}, as potential bottlenecks. Hardware experiments show comparable performance \textbf{while mitigating the reconstruction attacks from Section~\ref{sec:security-vulnerabilities}.}

\begin{table}[t]
\centering
\caption{
Summary of selected CILC benchmarking results and parameter configurations most relevant to our discussion (Section~\ref{sec:discussion}).
The three highlighted protocols capture the primary tradeoffs: \mascot (\textit{malicious secure}, strongest security, highest \onlinephase overhead), \hemi (\textit{semi-honest secure}, lowest measured \onlinephase overhead among the tested SMPC protocols), and \baseline (lowest overhead, but insecure).
See Table~\ref{tab:score-functions} for \score definitions.
Note \textit{secure} and \textit{insecure} columns use different units.
}
\label{tab:results}
\resizebox{\linewidth}{!}{%
\begin{tabular}{c cl  rr rr  rr rr  rr rr }
\toprule
& & & \multicolumn{4}{c}{\makecell{\mascot \\[-1pt] {\scriptsize Malicious Secure}}} & \multicolumn{4}{c}{\makecell{\hemi \\[-1pt] {\scriptsize Semi-Honest Secure}}} & \multicolumn{4}{c}{\makecell{\baseline (\cpp) \\[-1pt] {\scriptsize Insecure}}} \\
\cmidrule(lr){4-7} \cmidrule(lr){8-11} \cmidrule(lr){12-15}
$N$ & ID & Score Name & \multicolumn{2}{c}{$t$ [ms] $\downarrow$} & \multicolumn{2}{c}{Comms} & \multicolumn{2}{c}{$t$ [ms]} & \multicolumn{2}{c}{Comms} & \multicolumn{2}{c}{$t$ [$\mu$s]} & \multicolumn{2}{c}{Comms} \\
\cmidrule(lr){4-5} \cmidrule(lr){6-7} \cmidrule(lr){8-9} \cmidrule(lr){10-11} \cmidrule(lr){12-13} \cmidrule(lr){14-15}
& & & mean & std & [MB] & [\#] & mean & std & [MB] & [\#] & mean & std & [KB] & [\#] \\
\midrule
\multirow{10}{*}{16}
& \Sone & \texttt{L1\_dist\_bar} &
52.32 & 3.32 & 3.266 & 311 &
19.33 & 2.16 & 3.264 & 208 &
 1.45 & 0.05 & 0.128 & 2 \\
& \Stwo & \texttt{L1\_dist} &
2.97 & 0.10 & 0.063 & 57 &
1.11 & 0.20 & 0.062 & 34 &
0.39 & 0.02 & 0.128 & 2 \\
& \Sthree & \texttt{L2\_dist\_hat} &
102.81 & 2.94 & 5.039 & 1465 &
32.30 & 0.70 & 5.029 & 1072 &
 1.80 & 0.07 & 0.128 & 2 \\
& \Sfour & \texttt{L2\_dist} &
21.25 & 1.36 & 0.355 & 623 &
 6.72 & 0.15 & 0.350 & 458 &
 0.50 & 0.02 & 0.128 & 2 \\
& \Sfive & \texttt{L2sq\_dist\_hat} &
82.83 & 1.98 & 4.688 & 871 &
26.25 & 1.48 & 4.681 & 628 &
 2.30 & 0.32 & 0.128 & 2 \\
& \Ssix & \texttt{L2sq\_dist} &
0.75 & 0.16 & 0.002 & 29 &
0.39 & 0.09 & 0.001 & 14 &
0.57 & 0.09 & 0.128 & 2 \\
& \Sseven & \texttt{cossim\_hat} &
115.35 & 3.93 & 5.521 & 1695 &
 35.76 & 1.02 & 5.507 & 1242 &
  1.98 & 0.39 & 0.128 & 2 \\
& \Seight & \texttt{cossim\_2sqrt} &
35.55 & 1.38 & 0.836 & 853 &
10.38 & 0.34 & 0.829 & 628 &
 0.75 & 0.15 & 0.128 & 2 \\
& \Snine & \texttt{cossim\_1sqrt} &
29.52 & 1.53 & 0.488 & 853 &
 8.89 & 0.19 & 0.482 & 628 &
 0.76 & 0.09 & 0.128 & 2 \\
& \Sten & \texttt{dotprod} &
0.70 & 0.15 & 0.002 & 29 &
0.40 & 0.09 & 0.001 & 14 &
0.52 & 0.22 & 0.128 & 2 \\
\midrule
256 & \Sten & \texttt{dotprod} &
1.22 & 0.14 & 0.013 & 29 &
0.52 & 0.09 & 0.009 & 14 &
1.72 & 0.56 & 2.048 & 2 \\
\midrule
512 & \Sten & \texttt{dotprod} &
1.53 & 0.13 & 0.026 & 29 &
0.62 & 0.10 & 0.017 & 14 &
3.25 & 0.45 & 4.096 & 2 \\
\midrule
768 & \Sten & \texttt{dotprod} &
1.80 & 0.18 & 0.038 & 29 &
0.79 & 0.12 & 0.025 & 14 &
6.26 & 1.72 & 6.144 & 2 \\
\midrule
1024 & \Sten & \texttt{dotprod} &
 2.23 & 0.15 & 0.050 & 29 &
 0.88 & 0.11 & 0.034 & 14 &
14.10 & 4.80 & 8.192 & 2 \\
\midrule
\multirow{4}{*}{1536}
& \Stwo & \texttt{L1\_dist} &
194.17 & 6.45 & 6.008 & 2657 &
58.94 & 1.83 & 5.972 & 1762 &
 8.89 & 0.08 & 12.288 & 2 \\
& \Sfour & \texttt{L2\_dist} &
27.52 & 1.74 & 0.501 & 629 &
8.78 & 0.26 & 0.447 & 458 &
17.05 & 6.65 & 12.288 & 2 \\
& \Ssix & \texttt{L2sq\_dist} &
 4.01 & 0.17 & 0.075 & 35 &
 2.50 & 0.50 & 0.050 & 14 &
19.39 & 8.09 & 12.288 & 2 \\
& \Sten & \texttt{dotprod} &
2.70 & 0.11 & 0.075 & 29 &
1.16 & 0.17 & 0.050 & 14 &
9.13 & 1.83 & 12.288 & 2 \\
\midrule
2048 & \Sten & \texttt{dotprod} &
 3.69 &  0.67 &  0.099 & 29 &
 1.33 &  0.09 &  0.066 & 14 &
23.23 & 22.64 & 16.384 & 2 \\
\midrule
4096 & \Sten & \texttt{dotprod} &
 7.90 & 2.11 &  0.198 & 29 &
 2.42 & 0.36 &  0.132 & 14 &
37.09 & 4.68 & 32.768 & 2 \\
\midrule
\multirow{4}{*}{8192}
& \Stwo & \texttt{L1\_dist} &
1043.92 & 19.76 & 32.043 & 14107 &
319.21 & 10.18 & 31.85 & 9378 &
50.06 & 1.74 & 65.536 & 2 \\
& \Sfour & \texttt{L2\_dist} &
41.72 & 2.19 & 1.139 & 657 &
17.48 & 0.27 & 0.873 & 458 &
71.88 & 5.22 & 65.536 & 2 \\
& \Ssix & \texttt{L2sq\_dist} &
19.86 & 5.42 & 0.394 & 63 &
10.28 & 0.26 & 0.263 & 14 &
51.44 & 7.53 & 65.536 & 2 \\
& \Sten & \texttt{dotprod} &
 9.42 & 0.91 &  0.394 & 29 &
 4.57 & 0.72 &  0.263 & 14 &
41.75 & 7.10 & 65.536 & 2 \\
\bottomrule
\end{tabular}
}
\end{table}

\section{Conclusion \& Future Work}
\label{sec:conclusion}

To the best of our knowledge, CILC is the first system to integrate SMPC directly into a multi-agent CSLAM pipeline.
Typical CSLAM frameworks implicitly assume trusted peers, so continuously broadcasting GD exposes observational information to compromised swarm agents. Protecting GD comparisons is therefore both consequential and unusually well-suited to SMPC.
By restricting SMPC to the computationally lightweight ILC candidate detection step, CILC demonstrates that online cryptographic security -- despite expected additional overhead -- can remain real-time and comms-feasible while mitigating information leakage to compromised swarm agents.

Several directions remain for future work. Scaling beyond the current hardware demonstration (2 agents in a small map) is a necessary next step. Comparing multiple GDs simultaneously could enable more efficient comparisons (e.g., via secure $k$-d trees) while simultaneous comparisons between 3+ agents allow the use of more efficient \textit{honest-majority} protocols. Given SMPC's substantial gains from hardware acceleration (FPGAs, ASICs), privacy-preserving hardware tailored to multi-agent robotics is also promising. Finally, reputation and trust mechanisms \cite{josang2007survey,cavorsi2023dynamic,cavorsi2024Exploiting} could be implemented under SMPC to incentivize honest inputs to CILC. More broadly, CILC illustrates that carefully targeting privacy-sensitive components of a robotics pipeline can make practical cryptographic protection feasible without securing the entire system.

\section*{Acknowledgements}

\noindent
Thanks to Sarah Scheffler, Emily Shen, Trevor Ashley, J.~Parker Diamond, Dan McGann, and Aneesa Sonawalla for their valuable feedback. Figure~\ref{fig:system-diagram} was created using base artwork generated by ChatGPT. Claude and ChatGPT were used for text-level refinements.

\bibliographystyle{IEEEtran}
\bibliography{IEEEabrv,mybib}{}

\begin{thebibliography}{10}
\providecommand{\url}[1]{#1}
\csname url@samestyle\endcsname
\providecommand{\newblock}{\relax}
\providecommand{\bibinfo}[2]{#2}
\providecommand{\BIBentrySTDinterwordspacing}{\spaceskip=0pt\relax}
\providecommand{\BIBentryALTinterwordstretchfactor}{4}
\providecommand{\BIBentryALTinterwordspacing}{\spaceskip=\fontdimen2\font plus
\BIBentryALTinterwordstretchfactor\fontdimen3\font minus \fontdimen4\font\relax}
\providecommand{\BIBforeignlanguage}[2]{{%
\expandafter\ifx\csname l@#1\endcsname\relax
\typeout{** WARNING: IEEEtran.bst: No hyphenation pattern has been}%
\typeout{** loaded for the language `#1'. Using the pattern for}%
\typeout{** the default language instead.}%
\else
\language=\csname l@#1\endcsname
\fi
#2}}
\providecommand{\BIBdecl}{\relax}
\BIBdecl

\bibitem{dorigo2021swarm}
M.~Dorigo, G.~Theraulaz, and V.~Trianni, ``Swarm {{Robotics}}: {{Past}}, {{Present}}, and {{Future}} [{{Point}} of {{View}}],'' \emph{Proceedings of the IEEE}, vol. 109, no.~7, pp. 1152--1165, Jul. 2021.

\bibitem{ackerman2025unitree}
\BIBentryALTinterwordspacing
E.~Ackerman, ``Exploit allows for takeover of fleets of {Unitree} robots,'' \emph{{IEEE} Spectrum}, 9 2025, updated 29 Sep 2025. Published in print as ``{UniPwn} Exploit Allowed for Takeover of {Unitree} Robots,'' \textit{IEEE Spectrum}, Dec.\ 2025. [Online]. Available: \url{https://spectrum.ieee.org/unitree-robot-exploit}
\BIBentrySTDinterwordspacing

\bibitem{przymus2025wolves}
P.~Przymus and T.~Durieux, ``Wolves in the repository: A software engineering analysis of the xz utils supply chain attack,'' in \emph{IEEE/ACM 22nd Int.\ Conference on Mining Software Repositories (MSR)}.\hskip 1em plus 0.5em minus 0.4em\relax IEEE, 2025, pp. 91--102.

\bibitem{choudhary2017distributed}
S.~Choudhary \emph{et~al.}, ``Distributed mapping with privacy and communication constraints: {{Lightweight}} algorithms and object-based models,'' \emph{The International Journal of Robotics Research}, vol.~36, no.~12, pp. 1286--1311, Oct. 2017.

\bibitem{tian2020asynchronous}
Y.~Tian \emph{et~al.}, ``Asynchronous and {{Parallel Distributed Pose Graph Optimization}},'' \emph{IEEE Robotics Autom.\ Lett.}, vol.~5, no.~4, pp. 5819--5826, Oct. 2020.

\bibitem{tian2021distributed}
------, ``Distributed {{Certifiably Correct Pose-Graph Optimization}},'' \emph{IEEE Trans.\ on Robotics}, vol.~37, no.~6, pp. 2137--2156, Dec. 2021.

\bibitem{chang2021kimeramulti}
Y.~Chang \emph{et~al.}, ``Kimera-{{Multi}}: A {{System}} for {{Distributed Multi-Robot Metric-Semantic Simultaneous Localization}} and {{Mapping}},'' in \emph{{{IEEE Int.\ Conf.}} on {{Robotics}} and {{Automation}} ({{ICRA}})}, May 2021, pp. 11\,210--11\,218.

\bibitem{tian2022kimeramulti}
Y.~Tian \emph{et~al.}, ``Kimera-{{Multi}}: {{Robust}}, {{Distributed}}, {{Dense Metric-Semantic SLAM}} for {{Multi-Robot Systems}},'' \emph{IEEE Trans.\ on Robotics}, vol.~38, no.~4, pp. 2022--2038, Aug. 2022.

\bibitem{cieslewski2017efficient}
T.~Cieslewski and D.~Scaramuzza, ``Efficient {{Decentralized Visual Place Recognition Using}} a {{Distributed Inverted Index}},'' \emph{IEEE Robotics Autom.\ Lett.}, vol. 2(2), pp. 640--647, Apr. 2017.

\bibitem{bottaCyberSecurityRobots2023}
A.~Botta \emph{et~al.}, ``Cyber security of robots: {{A}} comprehensive survey,'' \emph{Intelligent Systems with Applications}, vol.~18, p. 200237, May 2023.

\bibitem{dwork2014algorithmic}
C.~Dwork and A.~Roth, ``The {{Algorithmic Foundations}} of {{Differential Privacy}},'' \emph{Foundations and Trends\textregistered{} in Theoretical Computer Science}, vol.~9, no. 3-4, pp. 211--487, Aug. 2014.

\bibitem{leny2014differentially}
J.~Le~Ny and G.~J. Pappas, ``Differentially {{Private Filtering}},'' \emph{IEEE Trans.\ on Automatic Control}, vol.~59, no.~2, pp. 341--354, Feb. 2014.

\bibitem{cortes2016differential}
J.~Cort{\'e}s \emph{et~al.}, ``Differential privacy in control and network systems,'' in \emph{{{IEEE}} 55th {{Conference}} on {{Decision}} and {{Control}} ({{CDC}})}, Dec. 2016, pp. 4252--4272.

\bibitem{han2017differentially}
S.~Han, U.~Topcu, and G.~J. Pappas, ``Differentially {{Private Distributed Constrained Optimization}},'' \emph{IEEE Trans.\ on Automatic Control}, vol.~62, no.~1, pp. 50--64, Jan. 2017.

\bibitem{hanPrivacyControlDynamical2018a}
S.~Han and G.~J. Pappas, ``Privacy in {{Control}} and {{Dynamical Systems}},'' \emph{Annual Review of Control, Robotics, and Autonomous Systems}, vol.~1, no.~1, pp. 309--332, May 2018.

\bibitem{benvenutiGuaranteedFeasibilityDifferentially2024}
A.~Benvenuti \emph{et~al.}, ``Guaranteed {{Feasibility}} in {{Differentially Private Linearly Constrained Convex Optimization}},'' \emph{IEEE Control Systems Letters}, vol.~8, pp. 2745--2750, 2024.

\bibitem{shoukry2016privacyaware}
Y.~Shoukry \emph{et~al.}, ``Privacy-aware quadratic optimization using partially homomorphic encryption,'' in \emph{{{IEEE}} {{Conference}} on {{Decision}} and {{Control}} ({{CDC}})}, Dec. 2016, pp. 5053--5058.

\bibitem{evans2018pragmatic}
D.~Evans, V.~Kolesnikov, and M.~Rosulek, ``A {{Pragmatic Introduction}} to {{Secure Multi-Party Computation}},'' \emph{Foundations and Trends\textregistered{} in Privacy and Security}, vol.~2, no. 2-3, pp. 70--246, Dec. 2018.

\bibitem{shi2020consensus}
\BIBentryALTinterwordspacing
E.~Shi, ``{Foundations of Distributed Consensus and Blockchains},'' 2020, book manuscript. [Online]. Available: \url{https://www.distributedconsensus.net}
\BIBentrySTDinterwordspacing

\bibitem{josang2007survey}
A.~J{\o}sang, R.~Ismail, and C.~Boyd, ``A survey of trust and reputation systems for online service provision,'' \emph{Decision Support Systems}, vol.~43, no.~2, pp. 618--644, Mar. 2007.

\bibitem{hastings2019sok}
M.~Hastings \emph{et~al.}, ``{{SoK}}: {{General Purpose Compilers}} for {{Secure Multi-Party Computation}},'' in \emph{2019 {{IEEE Symposium}} on {{Security}} and {{Privacy}} ({{SP}})}, May 2019, pp. 1220--1237.

\bibitem{keller2020mpspdz}
M.~Keller, ``{{MP-SPDZ}}: {{A Versatile Framework}} for {{Multi-Party Computation}},'' in \emph{Proceedings of the 2020 {{ACM SIGSAC Conference}} on {{Computer}} and {{Communications Security}}}.\hskip 1em plus 0.5em minus 0.4em\relax Virtual Event USA: ACM, Oct. 2020, pp. 1575--1590.

\bibitem{alexandru2020secure}
A.~B. Alexandru and G.~J. Pappas, ``Secure {{Multi-party Computation}} for {{Cloud-Based Control}},'' in \emph{Privacy in {{Dynamical Systems}}}, F.~Farokhi, Ed.\hskip 1em plus 0.5em minus 0.4em\relax Springer, 2020, pp. 179--207.

\bibitem{kammSecureFloatingPoint2015}
L.~Kamm and J.~Willemson, ``Secure floating point arithmetic and private satellite collision analysis,'' \emph{International Journal of Information Security}, vol.~14, no.~6, pp. 531--548, Nov. 2015.

\bibitem{alsayegh2022PrivacyPreserving}
M.~Alsayegh \emph{et~al.}, ``Privacy-{{Preserving Multi-Robot Task Allocation}} via {{Secure Multi-Party Computation}},'' in \emph{{{European Control Conference}} ({{ECC}})}, Jul. 2022, pp. 1274--1281.

\bibitem{diamond2025optimizing}
J.~P. Diamond \emph{et~al.}, ``Optimizing {{Local Computation}} in {{Secure Matrix Multiplication}} for {{Outsourced Neural Networks}},'' in \emph{2025 {{IEEE High Performance Extreme Computing Conference}} ({{HPEC}})}, Sep. 2025, pp. 1--7.

\bibitem{choncholasSnailSecureSingle2024}
J.~Choncholas \emph{et~al.}, ``Snail: {{Secure Single Iteration Localization}},'' \emph{arXiv preprint arXiv:2403.14916}, Mar. 2024.

\bibitem{speciale2019privacy}
P.~Speciale \emph{et~al.}, ``Privacy {{Preserving Image-Based Localization}},'' in \emph{2019 {{IEEE}}/{{CVF Conference}} on {{Computer Vision}} and {{Pattern Recognition}} ({{CVPR}})}.\hskip 1em plus 0.5em minus 0.4em\relax Long Beach, CA, USA: IEEE, Jun. 2019, pp. 5488--5498.

\bibitem{speciale2019privacya}
------, ``Privacy {{Preserving Image Queries}} for {{Camera Localization}},'' in \emph{2019 {{IEEE}}/{{CVF International Conference}} on {{Computer Vision}} ({{ICCV}})}.\hskip 1em plus 0.5em minus 0.4em\relax Seoul, Korea (South): IEEE, Oct. 2019, pp. 1486--1496.

\bibitem{geppert2020privacy}
M.~Geppert \emph{et~al.}, ``Privacy {{Preserving Structure-from-Motion}},'' in \emph{Computer {{Vision}} -- {{ECCV}} 2020}, A.~Vedaldi \emph{et~al.}, Eds.\hskip 1em plus 0.5em minus 0.4em\relax Cham: Springer Int.\ Pub., 2020, vol. 12346, pp. 333--350.

\bibitem{dusmanu2021privacypreserving}
M.~Dusmanu \emph{et~al.}, ``Privacy-{{Preserving Image Features}} via {{Adversarial Affine Subspace Embeddings}},'' in \emph{{{IEEE}}/{{CVF Conference}} on {{Computer Vision}} and {{Pattern Recognition}} ({{CVPR}})}.\hskip 1em plus 0.5em minus 0.4em\relax IEEE, Jun. 2021, pp. 14\,262--14\,272.

\bibitem{geppert2021privacy}
M.~Geppert \emph{et~al.}, ``Privacy {{Preserving Localization}} and {{Mapping}} from {{Uncalibrated Cameras}},'' in \emph{{{IEEE}}/{{CVF Conference}} on {{Computer Vision}} and {{Pattern Recognition}} ({{CVPR}})}.\hskip 1em plus 0.5em minus 0.4em\relax IEEE, Jun. 2021, pp. 1809--1819.

\bibitem{geppert2022privacy}
------, ``Privacy {{Preserving Partial Localization}},'' in \emph{{{IEEE}}/{{CVF Conference}} on {{Computer Vision}} and {{Pattern Recognition}} ({{CVPR}})}.\hskip 1em plus 0.5em minus 0.4em\relax IEEE, Jun. 2022, pp. 17\,316--17\,326.

\bibitem{pan2023privacy}
L.~Pan \emph{et~al.}, ``Privacy {{Preserving Localization}} via {{Coordinate Permutations}},'' in \emph{{{IEEE}}/{{CVF Int. Conf.}} on {{Computer Vision}} ({{ICCV}})}.\hskip 1em plus 0.5em minus 0.4em\relax IEEE, Oct. 2023, pp. 18\,128--18\,137.

\bibitem{galvezlopezDBoW2012}
D.~{Galvez-L{\'o}pez} and J.~D. Tardos, ``Bags of {{Binary Words}} for {{Fast Place Recognition}} in {{Image Sequences}},'' \emph{IEEE Trans.\ on Robotics}, vol.~28, no.~5, pp. 1188--1197, Oct. 2012.

\bibitem{arandjelovic2016netvlad}
R.~Arandjelovic \emph{et~al.}, ``{{NetVLAD}}: {{CNN Architecture}} for {{Weakly Supervised Place Recognition}},'' \emph{Proceedings of the IEEE conference on computer vision and pattern recognition}, pp. 5297--5307, 2016.

\bibitem{oquab2024dinov2}
M.~Oquab \emph{et~al.}, ``{{DINOv2}}: {{Learning Robust Visual Features}} without {{Supervision}},'' Feb. 2024.

\bibitem{simeoni2025dinov3}
O.~Sim{\'e}oni \emph{et~al.}, ``{{DINOv3}},'' Aug. 2025.

\bibitem{chelani2021how}
K.~Chelani, F.~Kahl, and T.~Sattler, ``How {{Privacy-Preserving}} are {{Line Clouds}}? {{Recovering Scene Details}} from {{3D Lines}},'' in \emph{{{IEEE}}/{{CVF Conference}} on {{Computer Vision}} and {{Pattern Recognition}} ({{CVPR}})}.\hskip 1em plus 0.5em minus 0.4em\relax IEEE, Jun. 2021, pp. 15\,663--15\,673.

\bibitem{carlone2025slam}
L.~Carlone \emph{et~al.}, \emph{{{SLAM Handbook}}: {{From Localization}} and {{Mapping}} to {{Spatial Intelligence}}}.\hskip 1em plus 0.5em minus 0.4em\relax Camb.\ Uni.\ Press, Oct. 2025.

\bibitem{murartalORBSLAM2015}
R.~{Mur-Artal}, J.~M.~M. Montiel, and J.~D. Tard{\'o}s, ``{{ORB-SLAM}}: {{A Versatile}} and {{Accurate Monocular SLAM System}},'' \emph{IEEE Trans.\ on Robotics}, vol. 31(5), pp. 1147--1163, Oct. 2015.

\bibitem{kim2018scan}
G.~Kim and A.~Kim, ``Scan {{Context}}: {{Egocentric Spatial Descriptor}} for {{Place Recognition Within 3D Point Cloud Map}},'' in \emph{2018 {{IEEE}}/{{RSJ International Conference}} on {{Intelligent Robots}} and {{Systems}} ({{IROS}})}, Oct. 2018, pp. 4802--4809.

\bibitem{deng2009imagenet}
J.~Deng \emph{et~al.}, ``Imagenet: {{A}} large-scale hierarchical image database,'' in \emph{IEEE Conf. on Computer Vision and Pattern Recognition}, 2009, pp. 248--255.

\bibitem{zhang2018unreasonable}
R.~Zhang \emph{et~al.}, ``The {{Unreasonable Effectiveness}} of {{Deep Features}} as a {{Perceptual Metric}},'' in \emph{2018 {{IEEE}}/{{CVF Conference}} on {{Computer Vision}} and {{Pattern Recognition}}}.\hskip 1em plus 0.5em minus 0.4em\relax Salt Lake City, UT: IEEE, Jun. 2018, pp. 586--595.

\bibitem{cavorsi2023dynamic}
M.~Cavorsi \emph{et~al.}, ``Dynamic {{Crowd Vetting}}: {{Collaborative Detection}} of {{Malicious Robots}} in {{Dynamic Communication Networks}},'' in \emph{62nd {{IEEE Conference}} on {{Decision}} and {{Control}} ({{CDC}})}, Dec. 2023, pp. 7546--7553.

\bibitem{cavorsi2024Exploiting}
------, ``Exploiting {{Trust}} for {{Resilient Hypothesis Testing With Malicious Robots}},'' \emph{IEEE Trans.\ on Robotics}, vol.~40, pp. 3514--3536, 2024.

\end{thebibliography}

\clearpage

\end{document}